\documentclass[authoryear]{elsarticle}
\usepackage{subfigure}      % Support for small, `sub' figures and tables. Your choice of alternative may be preferred instead.
\usepackage{multicol}       % sezioni multicolonna
\usepackage{multirow}       % multiriga belle tabelle
\usepackage{rotating}
\usepackage{url}
\usepackage[colorlinks=true,linkcolor=blue,urlcolor=black,bookmarksopen=true]{hyperref}
\usepackage{dsfont,mathrsfs}
\usepackage[T1]{fontenc}
\usepackage{epsfig,graphicx,amsbsy,amsfonts,amssymb,amsmath}

\usepackage{tikz}
\usetikzlibrary{decorations.pathmorphing} % noisy shapes
\usetikzlibrary{fit}				      % fitting shapes to coordinates
\usetikzlibrary{backgrounds}	          % drawing the background after the foreground

\usepackage{array}
\newcolumntype{L}[1]{>{\raggedright\arraybackslash}p{#1}}
\newcolumntype{C}[1]{>{\centering\arraybackslash}p{#1}}
\newcolumntype{R}[1]{>{\raggedleft\arraybackslash}p{#1}}

\usepackage[a4paper,top=3cm,bottom=3cm,left=3cm,right=3cm]{geometry}

\begin{document}

%\jvol{00} \jnum{00} \jyear{2013} \jmonth{December}
%%\articletype{Article}

\title{A probabilistic network for the diagnosis of acute cardiopulmonary diseases}

\author[DISIA]{Alessandro Magrini}
\ead{magrini@disia.unifi.it}

\author[MARIONEGRI]{Davide Luciani} %\corref{cor1}}
\ead{davide.luciani@marionegri.it}

\author[DISIA]{Federico Mattia Stefanini}
\ead{stefanini@disia.unifi.it}

%\cortext[cor1]{Corresponding author.}

\address[DISIA]{Department of Statistics, Computer Science, Applications - University of Florence, Florence, Italy}
\address[MARIONEGRI]{IRCCS - Mario Negri Institute for Pharmacological Research, Milan, Italy}

\begin{abstract}
In this paper, the development of a probabilistic network
for the diagnosis of acute cardiopulmonary diseases is presented.
This paper is a draft version of the article published after peer review in 2018
%\cite{Magrini18}
(\href{https://doi.org/10.1002/bimj.201600206}{\url{https://doi.org/10.1002/bimj.201600206}}).
A panel of expert physicians collaborated to specify
the \textit{qualitative part},
that is a directed acyclic graph defining a factorization
of the joint probability distribution of domain variables.
The \textit{quantitative part}, that is the set of all
conditional probability distributions defined by each factor,
was estimated following the Bayesian paradigm: we applied an original
formal representation, characterized by a low number of
parameters and a parameterization intelligible for physicians,
elicited the joint prior distribution of parameters from
medical experts, and updated it by conditioning on a dataset
of hospital patient records using Markov Chain Monte Carlo simulation.
Refinement was iteratively performed until the probabilistic
network provided satisfactory Concordance Index values for
a selection of acute diseases and reasonable inference on
six fictitious patient cases.
The probabilistic network can be employed to perform medical
diagnosis on a total of 63 diseases (38 acute and 25 chronic)
on the basis of up to 167 patient findings.
%
%\noindent\textit{Classification codes}: 62F15, 62P10  
\end{abstract}


\maketitle

%%%%%%%%%%%%%%%%%%%%%%%%%%%%%%%%%%%%%%%%%%%%%%%%%%%%%%%%%%%%%%%%

\section{Introduction}
\label{sec:intro}

Medical diagnosis is the process of identifying the disease
a patient is affected by, based on the assessment of specific
risk factors, signs, symptoms and results of exams.
Probabilistic networks \citep{Koller09} are increasingly used
to support medical diagnosis, as they provide an efficient
representation of complex stochastic sy\-stems by exploiting
causal knowledge, and because several efficient algorithms
to perform probabilistic reasoning (evidence propagation)
are available \citep{Lucas04}.

Probabilistic networks are composed of a \textit{qualitative part},
that is a directed acyclic graph (DAG) defining a factorization
of the joint probability distribution of domain variables,
and of a \textit{quantitative part}, where each factor defines
a conditional probability distribution.
In medical problems, the DAG is often specified in terms
of causal relationships among variables according to
pathophysiological knowledge contained in the specialised
literature.
However, the information required to estimate the
quantitative part is typically scattered in many different
sources and varies greatly in quality \citep{Druzdzel00a}.
Medical literature represents the most reliable 
source of quantitative information, but it may
not cover all aspects of interest.
When this is the case, medical experts are a useful alternative
resource, even though their quantitative assessments may not
be reliable without special training \citep{Kahneman82}.
Clinical data from medical records are another valuable source of knowledge
to build a probabilistic network,
but they are typically limited to few variables
%like it happens in clinical trials,
or contain many missing values.
%like it is the case of hospital patient records.

In existing medical applications of probabilistic networks, the
quantitative part is typically estimated exploiting either expert knowledge
\citep{Nathwani97,Suojanen99,Andreassen91a,Diez97,Gaag02b,
Galan02,Lacave03,Charitos09,Luciani07,Leibovici07},
or a database of patient cases \citep{Middleton91,Wasyluk01}.
In this paper, we describe our experience in the
development of a probabilistic network for the
diagnosis of acute cardiopulmonary diseases,
where two sources of information, %quantitative
beliefs from medical experts and clinical data,
were exploited to estimate the quantitative part.
The probabilistic network was conceived as an extension of
BayPAD (\textit{Bayesian Pulmonary embolism Assisted Diagnosis}),
a probabilistic network for the dia\-gnosis of pulmonary embolism
\citep{Luciani07}.
The work involved a panel of medical experts from various specialties
and consisted of three stages.
In the first stage, the qualitative part was specified by
medical experts following the constraint DAG described in
\citep{Luciani12}.
In the se\-cond stage, we applied an original formal representation
to the quantitative part, characterized by a low number of
parameters and a parameterization intelligible for physicians,
and the joint prior distribution of parameters was elicited
from medical experts.
In the third stage, we updated the joint prior distribution
of parameters in the Bayesian paradigm by conditioning
on a dataset of hospital patient
records using Markov Chain Monte Carlo (MCMC) simulation.
The three stages were iterated until the probabilistic network
provided reasonable inference on six fictitious patient cases.
%In our approach, the dimensionality of the quantitative part 
%resulted less than one thousand parameters for more than
%$260$ variables, a feature significantly easing both
%elicitation of quantitative beliefs from medical experts
%and MCMC simulation.
The three stages were iterated until the probabilistic
network provided satisfactory Concordance Index values
for several acute diseases and reasonable inference
on several fictitious patient cases.

The paper is organized as follows.
In Section \ref{sec:dag_eli}, we provide details on
the specification of the qualitative part.
In Section \ref{sec:param_eli}, we explain the
formal repre\-sentation for the quantitative part
and the method to elicit the joint prior
distribution of parameters.
In Section \ref{sec:mcmc}, we present data and provide details
on Bayesian estimation of the quantitative part.
In Section \ref{sec:illus}, we illustrate the elicitation
task for two variables in the probabilistic network, and
compare the resulting prior di\-stribution with the
posterior distribution obtained from MCMC simulation.
In Section \ref{sec:refin}, we detail the refinement process.
Section \ref{sec:discuss} includes the discussion of our
contribution.

\section{Specification of the qualitative part}
\label{sec:dag_eli}

The qualitative part of a probabilistic network consists
of a directed acyclic graph (DAG) representing a factorization
of the joint probability distribution of variables.
Each node of the DAG represents a variable, that may receive
any number of directed edges, indicating on which variables
(parent variables) its probability distribution is conditioned.

\begin{figure}[!hbt]
\centering
% The state vector is represented by a blue circle.
% "minimum size" makes sure all circles have the same size
% independently of their contents.
\tikzstyle{state}=[circle,
                                    thick,
                                    minimum size=1.2cm,
                                    draw=black]                                    
                                    %draw=blue!80,
                                    %fill=blue!20]
% The measurement vector is represented by an orange circle.
\tikzstyle{measurement}=[circle,
                                                thick,
                                                minimum size=1.2cm,
                                                draw=black]                                                
                                                %draw=orange!80,
                                                %fill=orange!25]
% The control input vector is represented by a purple circle.
\tikzstyle{input}=[circle,
                                    thick,
                                    minimum size=1.2cm,
                                    draw=black]
                                    %draw=purple!40,
                                    %fill=purple!20]
% The input, state transition, and measurement matrices
% are represented by gray squares.
% They have a smaller minimal size for aesthetic reasons.
\tikzstyle{matrx}=[rectangle,
                                    thick,
                                    minimum size=1cm,
                                    draw=black]                                    
                                    %draw=gray!80,
                                    %fill=gray!20]
% The system and measurement noise are represented by yellow
% circles with a "noisy" uneven circumference.
% This requires the TikZ library "decorations.pathmorphing".
\tikzstyle{noise}=[circle,
                                    thick,
                                    minimum size=1.2cm,
                                    draw=black]
                                    %fill=yellow!40,
                                    %decorate,decoration={random steps,segment length=2pt,amplitude=2pt}]
% Everything is drawn on underlying gray rectangles with
% rounded corners.
\tikzstyle{background}=[rectangle,
                                                fill=gray!10,
                                                inner sep=0.2cm,
                                                rounded corners=5mm]
\begin{tikzpicture}[>=latex,text height=1.5ex,text depth=0.25ex]
    % "text height" and "text depth" are required to vertically
    % align the labels with and without indices.
  % The various elements are conveniently placed using a matrix:
  \matrix[row sep=0.5cm,column sep=0.5cm] {
    % First line: Control input
    &
        \node (u_k-1) [noise]{$\mathbf{V}_{R}$}; 
        &
        &
        %\node (u_k)   [input]{$\mathbf{u}_k$};     
        \\
	        % Second line: System noise & input matrix
       % \node (w_k-1) [noise] {$\mathbf{w}_{k-1}$}; 
       &
        \node (B_k-1) [measurement] {$\mathbf{V}_Q$}; %{$V_Q$};       %      
        &
        %\node (w_k)   [noise] {$\mathbf{w}_k$};     
        &
         %        \node (B_k)   [matrx] {$\mathbf{B}$};       
        &
         \\
        % Third line: State & state transition matrix
        %\node (A_k-2)                    
        &
        \node (x_k-1) [measurement] {$\mathbf{V}_{D}$}; &
        \node (A_k-1) [noise] {$\mathbf{V}_C$};       &
        %\node (x_k)   [state] {$\mathbf{x}_k$};      
        \\
        % Fourth line: Measurement noise & measurement matrix
        %\node (v_k-1) [noise] {$\mathbf{v}_{k-1}$}; 
        &
        \node (H_k-1) [measurement] {$\mathbf{V}_S$};       &
        %\node (v_k)   [noise] {$\mathbf{v}_k$};     
        &
        %\node (H_k)   [matrx] {$\mathbf{H}$};     
           \\
        % Fifth line: Measurement
        &
        %\node (z_k-1) [measurement] {$\mathbf{V}_{MC}$}; 
        \node (z_k-1) [noise] {$\mathbf{V}_{MC}$}; 
        &
        \node (zmo_k-1) [measurement] {$\mathbf{V}_{MO}$}; 
        &
        \node (zmm_k-1)   [noise] {$\mathbf{V}_{MM}$};     
        \\
    };
    % The diagram elements are now connected through arrows:
    \path[->]
        %(A_k-2) edge[thick] (x_k-1)	% The main path between the
         (A_k-1) edge[thick] (x_k-1)	% states via the state
         (A_k-1) edge[thick] (B_k-1)	% states via the state
         (A_k-1) edge[thick] (H_k-1)	% states via the state
        %(A_k-1) edge[thick] (x_k)		% transition matrices is
        %(x_k)   edge[thick] (A_k)		% accentuated.
      %  (A_k)   edge[thick] (x_k+1)	% x -> A -> x -> A -> ...
       % (x_k+1) edge[thick] (A_k+1)
        (x_k-1) edge[thick] (H_k-1)				% Output path x -> H -> z
        (H_k-1) edge[thick] (z_k-1)
        (H_k-1) edge[thick] (zmo_k-1)
        (H_k-1) edge[thick] (zmm_k-1)
        %(x_k)   edge (H_k)
        %(H_k)   edge (z_k)
        %(x_k+1) edge (H_k+1)
       % (H_k+1) edge (z_k+1)
        %(v_k-1) edge (z_k-1)				% Output noise v -> z
        %(v_k)   edge (z_k)
        %(v_k+1) edge (z_k+1)
        %(w_k-1) edge (x_k-1)				% System noise w -> x
        %(w_k)   edge (x_k)
        %(w_k+1) edge (x_k+1)
        (u_k-1) edge[thick] (B_k-1)				% Input path u -> B -> x
        (B_k-1) edge[thick] (x_k-1)
        %(u_k)   edge (B_k)
        %%(B_k)   edge (x_k)
        %(u_k+1) edge (B_k+1)
        %(B_k+1) edge (x_k+1)
        (A_k-1) edge[thick] (zmo_k-1)
        ;
    % Now that the diagram has been drawn, background rectangles
    % can be fitted to its elements. This requires the TikZ
    % libraries "fit" and "background".
    % Control input and measurement are labeled. These labels have
    % not been translated to English as "Measurement" instead of
    % "Messung" would not look good due to it being too long a word.
    \begin{pgfonlayer}{background}
      %  \node [background,
                    %fit=(u_k-1) (u_k+1),
      %              label=left:Entrance:] {};
      %  \node [background,
      %              fit=(w_k-1) (v_k-1) (A_k+1)] {};
      %  \node [background,
      %              fit=(z_k-1) (z_k+1),
      %              label=left:Measure:] {};
    \end{pgfonlayer}
\end{tikzpicture}
\caption{The maximal constraint DAG, where c-nodes are sets of medical variables:
$V_{R}$: aetiology;
$V_{C}$: epidemiology;
$V_{Q}$: pathogenesis.;
$V_{D}$: pathology;
$V_{S}$: pathophysiology;
$V_{MC}$: semiotics (patient's chief complaints);
$V_{MO}$: semiotics (future outcomes);
$V_{MM}$: semiotics (other manifestations).}
\label{fig:consdag}
\end{figure}
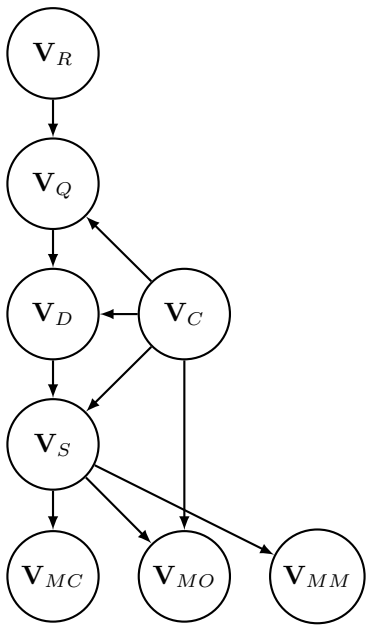

The qualitative part of our probabilistic network was specified
complying the constraint DAG (c-DAG, Figure \ref{fig:consdag})
described in \cite{Luciani12}, where c-nodes are sets of
variables and c-edges among c-nodes specify allowed directions 
of edges in the qualitative part of the probabilistic network.
Medical experts populated c-nodes of the c-DAG 
with relevant medical variables, as documented
in the specialised literature.
Edges that join nodes belonging to different c-nodes always
agree with c-edges, while eventual edges joining nodes belonging
to the same c-node were specified by medical experts,
without obeying any constraint besides the absence of directed
cycles in the resulting DAG.
The automated interview proposed by \cite{Luciani12}
was not adopted because it was conceived to derive the
DAG corresponding to a single patient presentation.

Due to the large number of variables, the resulting qualitative part
is not displayed, but the typology and the set of parent variables (parent set)
of each one is shown in Appendix \href{sec:appen2}.
Table \ref{tab:taxon} provides the classification of variables included
in the probabilistic network with respect to their statistical
and medical (as defined by the maximal constraint DAG) typology.
%
%It is worth noticing that no variables representing patient's chief
%complaints are included, because the qualitative part was specified
%by considering the most relevant diseases pertaining to the
%medical domain under analysis, instead of focusing on specific
%case reports.

\begin{table}[!bht]
\caption{Classification of variables included in the probabilistic network.
`Binary': categorical with one non-neutral category.
`Multi-valued': categorical with more than one non-neutral category.
No variables representing patient's chief complaints ($V_{MC}$) are
included, because the qualitative part was specified
by considering the most relevant diseases pertaining to the
medical domain under analysis, instead of focusing on specific
case reports.}
\label{tab:taxon}
\vspace{.3cm}
\centering \small
\begin{tabular}{l|rrrrrrrr|r}
  & $V_R$ & $V_C$ & $V_Q$ & $V_D$ & $V_S$ & $V_{MC}$ & $V_{MO}$ & $V_{MM}$ & \\ 
\hline
Binary       & 11 & 17 &  5 & 31 & 25 & 0 & 4 & 98 & 191\\ 
Multi-valued &  2 &  7 &  3 &  7 & 10 & 0 & 0 & 15 &  44\\ 
Continuous   &  0 &  1 &  0 &  0 & 14 & 0 & 0 & 12 &  27\\
\hline
             & 13 & 25 &  8 & 38 & 49 & 0 & 4 & 125 & 262 \\
\end{tabular}
\end{table}

%%%%%%%%%%%%%%%%%%%%%%%%%%%%%%%%%%%%%%%%%%%%%%%%%%%%%%%%%%%%%%%%

\section{Elicitation of the quantitative part}
\label{sec:param_eli}

The quantitative part of a probabilistic network
corresponds to the joint probability distribution
of domain variables, and it is factored
according to the DAG into the product of univariate
conditional distributions, one for each variable
given its parent variables in the DAG \citep{Koller09}.

The special case where each variable has finite sample
space is known as Bayesian network \citep{Korb10}.
The quantitative part of a Bayesian network is composed
of one Conditional Probability Table (CPT) for each variable.
%BayPAD is a Bayesian network.
In prominent medical applications of probabilistic networks
\citep{Nathwani97,Suojanen99,Andreassen91a,Diez97,Gaag02b,Wasyluk01,
Galan02,Lacave03,Charitos09,Luciani07,Leibovici07},
continuous variables underwent to discretization in order to
obtain a Bayesian network.
The major benefits of a Bayesian network consist of
a parameterization intelligible for a domain expert
(parameters are conditional probabilities)
and the availability of fast algorithms
for evidence propagation \citep{Yuan12}.
However, discretization of continuous variables
may dramatically increase the number of parameters
required to represent the quantitative part,
thus increasing uncertainty of estimates. 
We avoided discretization of continuous variables by
applying an original formal representation
to the quantitative part, characterized by a low number of
parameters and a parameterization intelligible for physicians.
In the proposed formal representation,
continuous variables are preliminarily rescaled
in order to assimilate the interpretation of quantitative
and qualitative values,
and a combination of the Beta regression and the
categorical logistic regression, reparameterized to help
physicians in performing quantitative assessments competently,
is exploited to model the distribution of each variable in the network.
The rescaling procedure and the two conditional
models are detailed in the remainder of this section.

\subsection{Rescaling procedure}
\label{sub:rescaling}

Medical categorical variables represent a qualitative
measure of a patient's condition, discriminating
among a healthy status and one or more pathological
conditions.
We assign value $0$ to the category associated
to a healthy patient condition and we refer to it
as the \textit{neutral value} of the variable.
Instead, we assign consecutive integer numbers
to the categories associated to pathological
patient conditions (non-neutral categories).
For example, a medical categorical variable with
a single non-neutral category will have sample
space $\{v:~v=0,1\}$, while the sample space will
be $\{v:~v=0,1,2\}$ if there are two non-neutral
categories.

The interpretation of measured values of medical
continuous variables is si\-milar, but it depends on
the scale of the variable at hand, thus medical
reasoning is more complicated.
The standard medical training and medical literature
provides physicians the ability to properly recognize the extreme
values of a medical continuous variable in a living patient,
as well as to distinguish among values involving normal
and pathological patient conditions \citep{Jacobs01,Irwin11}.
On these grounds, the scale $(v_{\hspace{.02cm}\scriptscriptstyle{\text{L2}}},v_{\hspace{.02cm}\scriptscriptstyle{\text{R2}}})$
of a medical continuous variable
$V$ is partitioned into three intervals:
\textit{n-range} $[v_{\hspace{.02cm}\scriptscriptstyle{\text{L1}}},v_{\hspace{.02cm}\scriptscriptstyle{\text{R1}}}]$,
in which va\-lues are regarded as non-pathological, 
\textit{lp-range} $(v_{\hspace{.02cm}\scriptscriptstyle{\text{L2}}},v_{\hspace{.02cm}\scriptscriptstyle{\text{L1}}})$,
including values lower than non-pathological ones,
and \textit{hp-range} $(v_{\hspace{.02cm}\scriptscriptstyle{\text{R1}}},v_{\hspace{.02cm}\scriptscriptstyle{\text{R2}}})$,
including values higher than non-pathological ones.
The mid value of \textit{n-range} is taken as the neutral value, while
%a reference for all values representing a fully healthy
%patient condition, and it is called \textit{neutral value}.
the mid values of \textit{lp-range} and \textit{hp-range}
are taken as reference for all values representing
hypo- or hyper-pathological conditions, respectively.
As a special case, one among \textit{lp-range} or
\textit{hp-range} may be of null size.

In order to make reasoning on quantitative values easier for physicians,
we propose a rescaling procedure making \textit{n-range},
\textit{lp-range} and \textit{hp-range} of equal size and
mapping their mid values to $0$, $-1$ and $1$, respectively:
\begin{equation}
  \widetilde{V} = \begin{cases}
  -1.5+\frac{V-v_{\hspace{.02cm}\scriptscriptstyle{\text{L2}}}}{v_{\hspace{.02cm}\scriptscriptstyle{\text{L1}}}-v_{\hspace{.02cm}\scriptscriptstyle{\text{L2}}}} &~~~\text{if } V< v_{\hspace{.02cm}\scriptscriptstyle{\text{L1}}}\\
  -0.5+\frac{V-v_{\hspace{.02cm}\scriptscriptstyle{\text{L1}}}}{v_{\hspace{.02cm}\scriptscriptstyle{\text{R1}}}-v_{\hspace{.02cm}\scriptscriptstyle{\text{L1}}}}              &~~~\text{if } v_{\hspace{.02cm}\scriptscriptstyle{\text{L1}}} \leq V < v_{\hspace{.02cm}\scriptscriptstyle{\text{R1}}}\\
  \phantom{+} 0.5+\frac{V-v_{\hspace{.02cm}\scriptscriptstyle{\text{R1}}}}{v_{\hspace{.02cm}\scriptscriptstyle{\text{R2}}}-v_{\hspace{.02cm}\scriptscriptstyle{\text{R1}}}}   &~~~\text{if } V \geq v_{\hspace{.02cm}\scriptscriptstyle{\text{R1}}}
  \end{cases}
\end{equation}
This way, a medical expert may refer to a value of a continuous
variable in terms of the relative position within
one among \textit{n-range}, \textit{lp-range} and \textit{hp-range},
instead of as a measured value on the original scale.
For instance, relative position $\varrho$ in \textit{lp-range}
corresponds to a rescaled value equal to $-1.5+\varrho$,
whereas relative position $\varrho$ in \textit{n-range} or
in \textit{hp-range} corresponds to a rescaled value equal
to $-0.5+\varrho$ and $0.5+\varrho$, respectively.
Furthermore, the rescaling procedure assimilates
the interpretation of quantitative
and qualitative medical scales, because
value $0$ of any variable represents a healthy
patient condition, and an unit variation from value $0$
is interpreted as a change of patient's state to
a reference pathological condition.

In the remainder, continuous variables are implicitly
considered as already rescaled.

\subsection{Categorical logistic regression}
\label{sub:catlogreg}

Consider a categorical variable included in the probabilistic
network, say $Y$, with $s_Y$ non-neutral values and
parent set $\boldsymbol{X}$.
All categorical variables in $\boldsymbol{X}$ with more
than one non-neutral value are replaced by a set of
dummy indicators, one for each non-neutral value,
obtaining the parent set $X_1,\ldots,X_n$.
The categorical logistic regression \citep[Chapter 5]{McCullagh89}
applied to $Y$ is:
\begin{equation}
\begin{gathered}
\text{log}\left(\frac{\text{Pr}(Y=y \mid x_1,\ldots,x_n)}{\text{Pr}(Y=0 \mid x_1,\ldots,x_n)}\right)=(1,x_1,\ldots,x_n)'\boldsymbol{\beta}^{(y)}\\
\boldsymbol{\beta}^{(y)}=(\beta_{0,y},\beta_{1,y},\ldots,\beta_{n,y}) \hspace{1.2cm} y=1,\ldots,s_Y\\
\end{gathered}
\end{equation}\label{eq:logisreg}
For each non-neutral value $y$ of $Y$,
parameters in $\boldsymbol{\beta}^{(y)}$
are regression coefficients on the logit scale
and are interpreted as log odds ratios:
\begin{equation}
\begin{gathered}
\beta_{0,y}=\text{log}\left(\frac{\pi_{0,y}}{\pi_{0,0}}\right) \hspace{1.2cm} y=1,\ldots,s_Y\\
\beta_{i,y}=\text{log}\left(\frac{\pi_{i,y}}{\pi_{i,0}}\right)-\beta_{0,y} \hspace{1.2cm} i=1,\ldots,n; ~y=1,\ldots,s_Y\\
\end{gathered}
\end{equation}
where:
\begin{equation}
\begin{gathered}
\pi_{0,y}=\text{Pr}[Y=y \mid X_1=0,\ldots,X_n=0] \hspace{1.2cm} y=0,1,\ldots,s_Y\\
\pi_{i,y}=\text{Pr}[Y=y \mid X_i=1,X_{j:j\not=i}=0] \hspace{1cm} i=1,\ldots,n; ~y=0,1,\ldots,s_Y
\end{gathered}
\end{equation}
If $X_i$ is a continuous parent variable, it holds:
\begin{equation}
\pi_{i,y}=\text{Pr}[Y=y \mid X_i=1,X_{j:j\not=i}=0]=1-\text{Pr}[Y=y \mid X_i=-1,X_{j:j\not=i}=0]
\end{equation}
The conditional probability of non-neutral value $y$ of $Y$
can be can be rewritten in terms of parameters
$\boldsymbol{\pi}^{(y)}=(\pi_{0,y},\pi_{1,y},\ldots,\pi_{n,y})$:
\begin{equation}\label{eq:cprob}
\begin{gathered}
\text{log}\left(\frac{\text{Pr}(Y=y \mid x_1,\ldots,x_n)}{\text{Pr}(Y=0 \mid x_1,\ldots,x_n)}\right)=\left(1-\sum_{i=1}^n x_i\right)\text{log}\left(\frac{\pi_{0,y}}{\pi_{0,0}}\right)+\sum_{i=1}^n x_i ~\text{log}\left(\frac{\pi_{i,y}}{\pi_{i,0}}\right)\\
y=1,\ldots,s_Y\\
\end{gathered}
\end{equation}
It follows that parameters are probabilities conditioned
to a configuration of parent variables where all but one
take value $0$, thus physicians
are expected to be competent in performing quantitative
assessments.
The prior distribution of $(\pi_{i,0},\pi_{i,1},\ldots,\pi_{i,s_Y})$
($i=0,1,\ldots,n$) is elicited using the Equivalent Prior Sample
(EPS) method \citep{Winkler67}:
\begin{equation}
(\pi_{i,0},\pi_{i,1},\ldots,\pi_{i,s_Y}) \sim \text{Dirichlet}\left(\hat{\pi}_{i,0}\cdot\hat{q}_{i,0},~\hat{\pi}_{i,1}\cdot\hat{q}_{i,1},\ldots,\hat{\pi}_{i,s_Y}\cdot\hat{q}_{i,s_Y}\right)\\
\end{equation}
where, for $y=0,1,\ldots,s_Y$, $\hat{\pi}_{i,y}$ is the assessment
of $\pi_{i,y}$ and $\hat{q}_{i,y}$ is the number of patient
cases on which $\hat{\pi}_{i,y}$ is based.

%The categorical logistic regression does not allow
%parameter $\pi_{i,y}$ to be $0$ or $1$ for any $i$ or $y$
%(Equation \ref{eq:cprob}), however a medical expert could
%assess $\pi_{i,y}=0$ or $\pi_{i,y}=1$ without uncertainty.
%In this case, we achieved an exact representation of the
%expert's belief up to the fourth decimal by replacing value
%$0$ with $10^{-5}$ and value $1$ with $1-10^{-5}$.

Value $0$ is not allowed for parameters $\pi_{0,0},\ldots,\pi_{0,s_Y}$.
This issue can be overcome by replacing zeros with a small number,
for example $10^{-7}$ to provide an exact representation up to
the sixth decimal.
Unfortunately, in this case or simply when
$\pi_{0,y}$ is small compared to $\pi_{i,y}$ for any $y>0$ and $i>0$,
the probability of the neutral value of $Y$ tends to become $0$ too fast
as more than one parent variable takes a non-neutral value.
Thus, we further refine the model in Equation \ref{eq:cprob}
as a mixture of two components:
a categorical probability distribution $\pi_{0,0},\ldots,\pi_{0,s_Y}$
when the sum of parent values is less or equal to zero,
and, otherwise, the model in Equation \ref{eq:cprob}
where parameters $\beta_{0,1},\ldots,\beta_{0,s_Y}$ are
replaced by new parameters $\eta_1,\ldots,\eta_{s_Y}$
unrelated to $\pi_{0,0},\ldots,\pi_{0,s_Y}$:
\begin{equation}
\begin{gathered}
\begin{cases}
\text{Pr}(Y=y \mid x_1,\ldots,x_n)=\pi_{0,y} & \text{if } \sum_{i=1}^n x_i \leq 0\\
\text{log}\left(\frac{\text{Pr}(Y=y \mid x_1,\ldots,x_n)}{\text{Pr}(Y=0 \mid x_1,\ldots,x_n)}\right)=
\left(1-\sum_{i=1}^n x_i\right)\eta_y+\sum_{i=1}^n x_i ~\text{log}\left(\frac{\pi_{i,y}}{\pi_{i,0}}\right) &  \text{otherwise}\\
\end{cases}\\
y=1,\ldots,s_Y
\end{gathered}\label{eq:logisreg3}
\end{equation}
For $i=1,\ldots,s_Y$, parameter $\eta_y$ is set equal to $0$ if $Y$
has less than two parent variables, otherwise a standard Gaussian prior
is assumed.

\subsection{Beta regression}
\label{sub:betareg}

Consider a continuous variable included in the probabilistic
network, say $Y$, with parent set $\boldsymbol{X}$.
All categorical variables in $\boldsymbol{X}$ with more
than one non-neutral value are replaced by a set of
dummy indicators, one for each non-neutral value,
obtaining the parent set $X_1,\ldots,X_n$.
The variable $\frac{Y+1.5}{3}$ has sample space $(0,1)$,
thus the Beta regression model \citep{Ferrari04} can
be applied:
\begin{equation}
\begin{gathered}
\frac{Y+1.5}{3} \sim \text{Beta}(\delta\tau,(1-\delta)\tau)\\
\text{log}\left(\frac{\delta}{1-\delta}\right)=\text{log}\left(\frac{\text{E}[Y \mid x_1,\ldots,x_n]+1.5}{1.5-\text{E}[Y \mid x_1,\ldots,x_n]}\right)=(1,x_1,\ldots,x_n)' \boldsymbol{\beta}
\end{gathered}
\end{equation}\label{eq:betareg}
Parameters $\boldsymbol{\beta}=(\beta_0,\beta_1,\ldots,\beta_n)$
are regression coefficients on the logit scale
and have the following interpretation:
\begin{equation}
\begin{gathered}
\beta_0=\text{log}\left(\frac{\mu_0+1.5}{1.5-\mu_0}\right)\\
\beta_i=\log\left(\frac{\mu_i+1.5}{1.5-\mu_i}\right)-\beta_0 \hspace{1.2cm} i=1,\ldots,n
\end{gathered}
\end{equation}
where:
\begin{equation}
\begin{gathered}
\mu_0=\text{E}[Y \mid X_1=0,\ldots,X_n=0]\\
\mu_i=\text{E}[Y \mid X_i=1,X_{j:j\not=i}=0] \hspace{1.2cm} i=1,\ldots,n
\end{gathered}
\end{equation}
If $X_i$ is a continuous parent variable, it holds:
\begin{equation}
\mu_i=\text{E}[Y \mid X_i=1,X_{j:j\not=i}=0]=-\text{E}[Y \mid X_i=-1,X_{j:j\not=i}=0]  %\hspace{.5cm} i=1,\ldots,n
\end{equation}
Parameter $\tau$ is constant across the configuration
of parent variables and regulates heteroscedasticity:
\begin{equation}
\text{Var}[Y \mid x_1,\ldots,x_n]=\frac{(\text{E}[Y \mid x_1,\ldots,x_n]+1.5)\cdot(1.5-\text{E}[Y \mid x_1,\ldots,x_n])}{(1+\tau)}
\end{equation}
The logit of the expected value of $Y$ can be
can be rewritten in terms of para\-meters
$\boldsymbol{\mu}=(\mu_0,\mu_1,\ldots,\mu_n)$:
\begin{equation}
\begin{gathered}
\log\left(\frac{\text{E}[Y \mid x_1,\ldots,x_n]+1.5}{1.5-\text{E}[Y \mid x_1,\ldots,x_n]}\right)=\left(1-\sum_{i=1}^n x_i\right)\log\left(\frac{\mu_0+1.5}{1.5-\mu_0}\right)+\\
  +\sum_{i=1}^n x_i~\log\left(\frac{\mu_i+1.5}{1.5-\mu_i}\right)
\end{gathered}\label{eq:cmean}
\end{equation}
This way, parameters are expected values conditioned
to a configuration of pa\-rent variables where all but one
take value $0$.
A medical expert is expected to be competent in performing
quantitative assessments under such parameterization,
because, thanks to the rescaling procedure,
he/she may refer to the expected value
of the response in terms of the relative position within
one among \textit{n-range}, \textit{lp-range} and \textit{hp-range}
(Subsection \ref{sub:rescaling}).
Typically, if no relevant parent variables are omitted,
the expected value of the response is $0$ when all parent
variables take value $0$, that is $\mu_0=0$ without
uncertainty, and equation \ref{eq:cmean} simplifies into:
\begin{equation}\label{eq:cmean2}
\log\left(\frac{\text{E}[Y \mid x_1,\ldots,x_n]+1.5}{1.5-\text{E}[Y \mid x_1,\ldots,x_n]}\right)=\sum_{i=1}^n x_i~\log\left(\frac{\mu_i+1.5}{1.5-\mu_i}\right)
\end{equation}
The prior distribution of parameter $\mu_i$ ($i=0,1,\ldots,n$)
is elicited using the Equivalent Prior Sample (EPS) method
\citep{Winkler67}:
\begin{equation}
\frac{\mu_i+1.5}{3} \sim \text{Beta}\left(\frac{\hat{\mu}_i+1.5}{3} ~\hat{q}_i, \left(1-\frac{\hat{\mu}_i+1.5}{3}\right)\hat{q}_i\right)\\
\end{equation}
where $\hat{\mu}_i$ is the assessment of $\mu_i$ and
$\hat{q}_i$ is the number of patient cases on which
$\hat{\mu}_i$ is based.
Typically, the expected value of $Y$ is equal to $0$
when all parent variables take value $0$,
that is parameter $\mu_0$ is equal to $0$ without uncertainty
and Equation \ref{eq:cmean} simplifies into:
\begin{equation}
\log\left(\frac{\text{E}[Y \mid x_1,\ldots,x_n]+1.5}{1.5-\text{E}[Y \mid x_1,\ldots,x_n]}\right)=\sum_{i=1}^n x_i~\log\left(\frac{\mu_i+1.5}{1.5-\mu_i}\right)
\end{equation}
A default prior distribution $\tau \sim \text{Gamma}(89.4917,2.0304)$
is assigned to the precision parameter,
implying a probability between $0.95$ and $0.99$
for the response to take value in \textit{n-range},
given that all parent variables take value $0$.

%%%%%%%%%%%%%%%%%%%%%%%%%%%%%%%%%%%%%%%%%%%%%%%%%%%%%%%%%%%%%%%%

\section{Bayesian estimation of the quantitative part}
\label{sec:mcmc}

Data collected by \cite{Squizzato11} were exploited to update
the joint prior distribution of parameters.
Let $\boldsymbol{\theta}$ be the set of parameters induced by
conditional models shown in Section \ref{sec:param_eli}.
Given a dataset of cases $\mathcal{D}$ from the problem domain,
the joint prior distribution $p(\boldsymbol{\theta})$
is updated in the Bayesian paradigm by computing
the joint posterior distribution
$p(\boldsymbol{\theta} \mid \mathcal{D}) \propto p(\mathcal{D} \mid \boldsymbol{\theta}) p(\boldsymbol{\theta})$.
Due to the large size of the probabilistic network,
direct computation of the joint posterior distribution
of parameters is intractable, therefore approximated
calculations were performed by means of Markov Chain Monte Carlo
(MCMC) simulation.

In the remainder of this section, we provide details
on data and MCMC implementation.

\subsection{Data}
\label{sub:data}

The clinical study described in \citep{Squizzato11} gathered a
total of $17497$ electronic admission records of patients referred
to an emergency department for cardiopulmonary disorders and then
hospitalised from January to June $2007$ in six italian hospitals.
A block random sampling design was applied to select $800$
hospitalised patients.
The randomization blocks were defined by the combination of age cate\-gories
(less than $30$, between $30$ and $60$, over $60$), gender and four
symptoms of pulmonary embolism: acute dyspnea, chest pain, fainting
and palpitations.
A further block was defined by the presence of either pulmonary
embolism, aortic dissection or pneumothorax.

Patients were selected within each block according to the probability
of the blocking variables in the whole population.
All patients hospitalised for trauma were excluded, leading to a total
of $750$ eligible records.
In our analysis, other $28$ patients were excluded because the
acute disease was not of cardiopulmonary origin ($23$ patients),
the visit in emergency department was planned in advance ($3$ patients),
or data were of poor quality ($2$ patients).
Baseline clinical characteristics of data are summarized in
Table \ref{tab:summ1}.

\begin{table}[!hbt]
\caption{Baseline clinical characteristics of data.}
\label{tab:summ1}
\centering \small
\vspace{.3cm}
\begin{tabular}{lr}
\hline
Age: median (interquartile range) & 68 (52-81)\phantom{0.}\\
Female                 & 342 (47.37\%)\\
Immobilisation         & 106 (14.68\%)\\
Surgery                &  24 \phantom{0}(3.32\%)\\
Fever                  &  81 (11.22\%)\\
\textit{Associated chronic diseases}  &\\
\hspace{.2cm} Aortic aneurysm                  & 41 \phantom{0}(5.68\%)\\
\hspace{.2cm} Cholelithiasis                   & 35 \phantom{0}(4.88\%)\\
\hspace{.2cm} Chronic obstructive pulmonary disease & 123 (17.04\%)\\
\hspace{.2cm} Chronic cerebro-vascular disease & 5 \phantom{0}(0.69\%)\\
\hspace{.2cm} Chronic venous insufficiency     & 69 \phantom{0}(9.56\%)\\
\hspace{.2cm} Neoplastic disease               & 108 (14.96\%)\\
\hspace{.2cm} Neuromuscular disease            & 6 \phantom{0}(0.83\%)\\
\hspace{.2cm} Thyroid disease                  & 7 \phantom{0}(0.97\%)\\
\hline
\end{tabular}
\end{table}

Since hospital patient records are not collected for medical purposes,
varia\-bles in the categories $V_Q$ (pathogenesis), $V_D$ (pathology)
and $V_S$ (pathophysiology) are typically unobserved,
and variables in the other categories may have not been reported
(missing), either because not of interest or obvious for the physician.
In order to reduce the number of missing values,
medical experts performed judgements on some unobserved variables
on the basis of the diagnosis yielded at hospital discharge,
and set explicit criteria to establish which missing data
could be safely assumed as neutral values.
The frequency distribution of each variable including
the percentage of missing values is shown in Appendix \href{sec:appen2}.

Overall, 66 variables (25\%) included in the probabilistic
network resulted completely unobserved in our dataset,
observed variables contained a total of 245 missing values (34\%),
for a total of 383 missing values (53\%) among all the variables
included in the probabilistic network.
Missing values typically arise when either a diagnosis is
not available, or the physician decides
that a certain ascertainment is not necessary.
In the former case, a datum is missing because
the observed ones are deemed insufficient by the
physician in order to formulate a diagnosis.
In the latter case, a datum is missing because the physician
believes that a certain ascertainment is irrelevant given
of the observed ones.
As such, missing values comply with the Missing at Random
assumption (MAR, \cite{Raghunathan04}).

\subsection{Markov Chain Monte Carlo implementation}
\label{sub:mcmc}

A sample from the joint posterior distribution of parameters
was obtained using MCMC algorithms available in
$\mathsf{JAGS}$ \citep{Plummer03}.
According to the MAR assumption, we ignored the mechanism generating missing values,
thus they were treated as unknown variables on the same footing of parameters.

The simulation was run for $55000$ iterations, of which the first
$30000$ were discarded and the others were thinned by an interval
of $5$ to reduce sample autocorrelation.
We applied the Geweke \citep{Geweke92}, Heidelberger-Welch \citep{Heidelberger83}
and Raftery-Lewis \citep{Raftery95} diagnostic tests to detect lack of convergence.
We obtained that less than $1\%$ of parameters passed no tests, more
than $99\%$ of parameters passed at least one test, more than $90\%$
of parameters passed at lest two tests and almost half of the parameters
passed all the three tests.

At the end of MCMC simulation, the divergence between the prior and the posterior
distribution of each parameter $\theta \in \boldsymbol{\theta}$
was quantified using the following statistic, that we call D-statistic:
\begin{equation}
\text{D}(\theta)=\frac{1}{S} \sum_{i=1}^S I^{(i)}(\theta)
%I_{E}(\theta)
%\mathds{1}_{\theta^*_i \in \hspace{.05cm} \text{I}(\theta; 0.95)}
\end{equation}
where $I^{(i)}(\theta)$ is a dummy indicator taking value $1$
if the $i$-th value of $\theta$ from MCMC simulation is
included in the equal-tail $95\%$ prior credible interval for $\theta$,
and $S$ is the length of the MCMC sample.
A value of the D-statistic above $0.95$ suggests a substantial agreement
between the medical experts' belief and data.
Values of the D-statistic near $0.95$ mean that the medical experts' belief
is not updated by data, provided that the posterior distribution
is unimodal.
Values of the D-statistic below $0.95$ indicate an increasing disagreement
between the medical experts' belief and data.
The maximum disagreement holds for values of the D-statistic near $0$,
meaning that the majority of posterior samples is outside the prior
$95\%$ credible interval.

%%%%%%%%%%%%%%%%%%%%%%%%%%%%%%%%%%%%%%%%%%%%%%%%%%%%%%%%%%%%%%%%%%%%%%%%%%%%

\section{Illustration}
\label{sec:illus}

In this section, we illustrate the elicitation
task for two variables in the proba\-bilistic network:
`Bradycardia/Tachycardia' and `Heart rate',
and compare the resulting marginal prior distributions with the
marginal posterior distributions obtained from MCMC simulation.

\subsection{Bradycardia/Tachycardia}

`Bradycardia/Tachycardia' is a categorical variable with sample space
\{`absent', `bradycardia', `moderate tachycardia', `severe tachycardia'\}
representing the absence or the presence of
bradycardia and/or tachycardia in a patient.
Its parent variables are `Heart drive',
hyper-restricted continuous variable here indicated as $X_1$, and
`Dehydration', hyper-restricted continuous variable here indicated as $X_2$.

\begin{figure}[!hbt]
\center
\includegraphics[width=1\columnwidth]{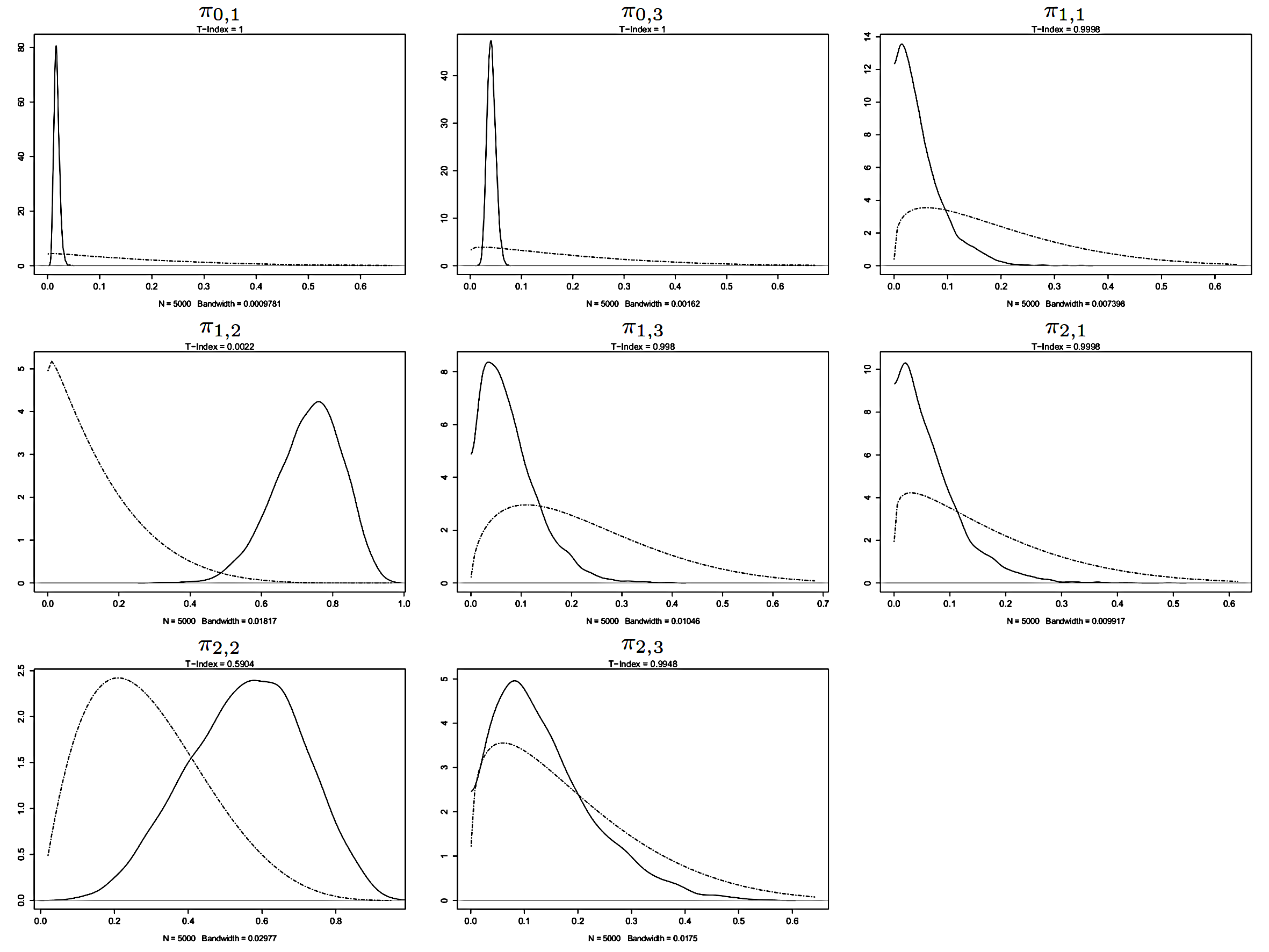}
\caption{Marginal probability density of parameters defining the conditional model of `Bradycardia/Tachycardia'.
Straight lines: posterior probability density. Dotted lines: prior probability density.
The D-statistic value is shown below the title.}
\label{fig:BT_mcmc}
\end{figure}

The probability distribution of `Bradycardia/Tachycardia' was
defined by applying the Beta regression model explained in
Subsection \ref{sub:catlogreg}:
The elicited prior distribution of parameters was:
\begin{equation}
\begin{gathered}
(\pi_{0,0}, \pi_{0,1}, \pi_{0,2}, \pi_{0,3}) \sim \text{Dirichlet}(3.88,1.03,0.00,1.09)\\          
(\pi_{1,0}, \pi_{1,1}, \pi_{1,2}, \pi_{1,3}) \sim \text{Dirichlet}(3.10,1.30,1.04,1.55)\\   
(\pi_{2,0}, \pi_{2,1}, \pi_{2,2}, \pi_{2,3}) \sim \text{Dirichlet}(2.50,1.15,2.05,1.30) 
\end{gathered}
\end{equation}
In Figure \ref{fig:BT_mcmc}, marginal prior distribution and 
kernel density estimate of marginal posterior distribution
of model parameters are shown.
The summary of marginal prior and posterior distribution of
parameters is provided in Table \ref{tab:BT_mcmc}.

\begin{table}%[!hbt]
\caption{Summary of marginal prior and posterior probability density of
parameters defining the conditional model of `Bradycardia/Tachycardia'.
`Std. dev.': standard deviation. `95\% QI': 95\% quantile interval.}
\label{tab:BT_mcmc}
\centering \footnotesize
\vspace{.3cm}
\begin{tabular}{lrrrrrr}
 & Prior & Prior     & \multirow{2}{*}{Prior 95\% QI} & Posterior & Posterior & \multirow{2}{*}{Posterior 95\% QI}\\
 & mean  & std. dev. &                                & mean      & std. dev. &\\
\hline
$\pi_{0,0}$ & 0.6467 & 0.0326 & (0.2647, 0.9383) & 0.9408 & 0.0098 & (0.9211, 0.9589)\\
$\pi_{0,1}$ & 0.1717 & 0.0203 & (0.0057, 0.5290) & 0.0178 & 0.0052 & (0.0089, 0.0293)\\
$\pi_{0,2}$ & 0.0000 & 0.0000 & (0.0000, 0.0000) & 0.0000 & 0.0000 & (0.0000, 0.0000)\\
$\pi_{0,3}$ & 0.1817 & 0.0212 & (0.0072, 0.5429) & 0.0414 & 0.0084 & (0.0263, 0.0588)\\
$\pi_{1,0}$ & 0.4435 & 0.0309 & (0.1275, 0.7897) & 0.1394 & 0.0695 & (0.0300, 0.2954)\\
$\pi_{1,1}$ & 0.1860 & 0.0189 & (0.0116, 0.5215) & 0.0512 & 0.0431 & (0.0025, 0.1615)\\
$\pi_{1,2}$ & 0.1488 & 0.0159 & (0.0050, 0.4685) & 0.7298 & 0.0933 & (0.5222, 0.8918)\\
$\pi_{1,3}$ & 0.2217 & 0.0216 & (0.0207, 0.5676) & 0.0795 & 0.0575 & (0.0071, 0.2275)\\
$\pi_{2,0}$ & 0.3571 & 0.0287 & (0.0768, 0.7136) & 0.2416 & 0.1293 & (0.0479, 0.5401)\\
$\pi_{2,1}$ & 0.1643 & 0.0172 & (0.0074, 0.4910) & 0.0678 & 0.0561 & (0.0031, 0.2015)\\
$\pi_{2,2}$ & 0.2929 & 0.0259 & (0.0462, 0.6489) & 0.5534 & 0.1538 & (0.2314, 0.8278)\\
$\pi_{2,3}$ & 0.1857 & 0.0189 & (0.0116, 0.5208) & 0.1372 & 0.0929 & (0.0134, 0.3565)\\
\hline
\end{tabular}
\end{table}

\subsection{Heart rate}

`Heart rate' is a continuous variable  measuring the heart rate in a patient.
Its parent variables are `Autonomic nervous system status', a categorical variable
with sample space \{`regular', `moderate adrenergic status', `severe adrenergic status',
`hypertensive crisis', `moderate cholinergic status', `severe cholinergic status'\},
representing the status of autonomic nervous system in a patient,
and `Bradycardia/Tachycardia', described above.
Since both parents are catego\-rical variables with more than one
non-neutral category, they are replaced by dummy indicators:
$X_1$, $X_2$, $X_3$, $X_4$ and $X_5$ represent the non-neutral categories of `Heart rate',
while $X_6$, $X_7$ and $X_8$ represent the non-neutral categories of `Bradycardia/Tachycardia'.

\begin{figure}[!h]
\center
\includegraphics[width=1\columnwidth]{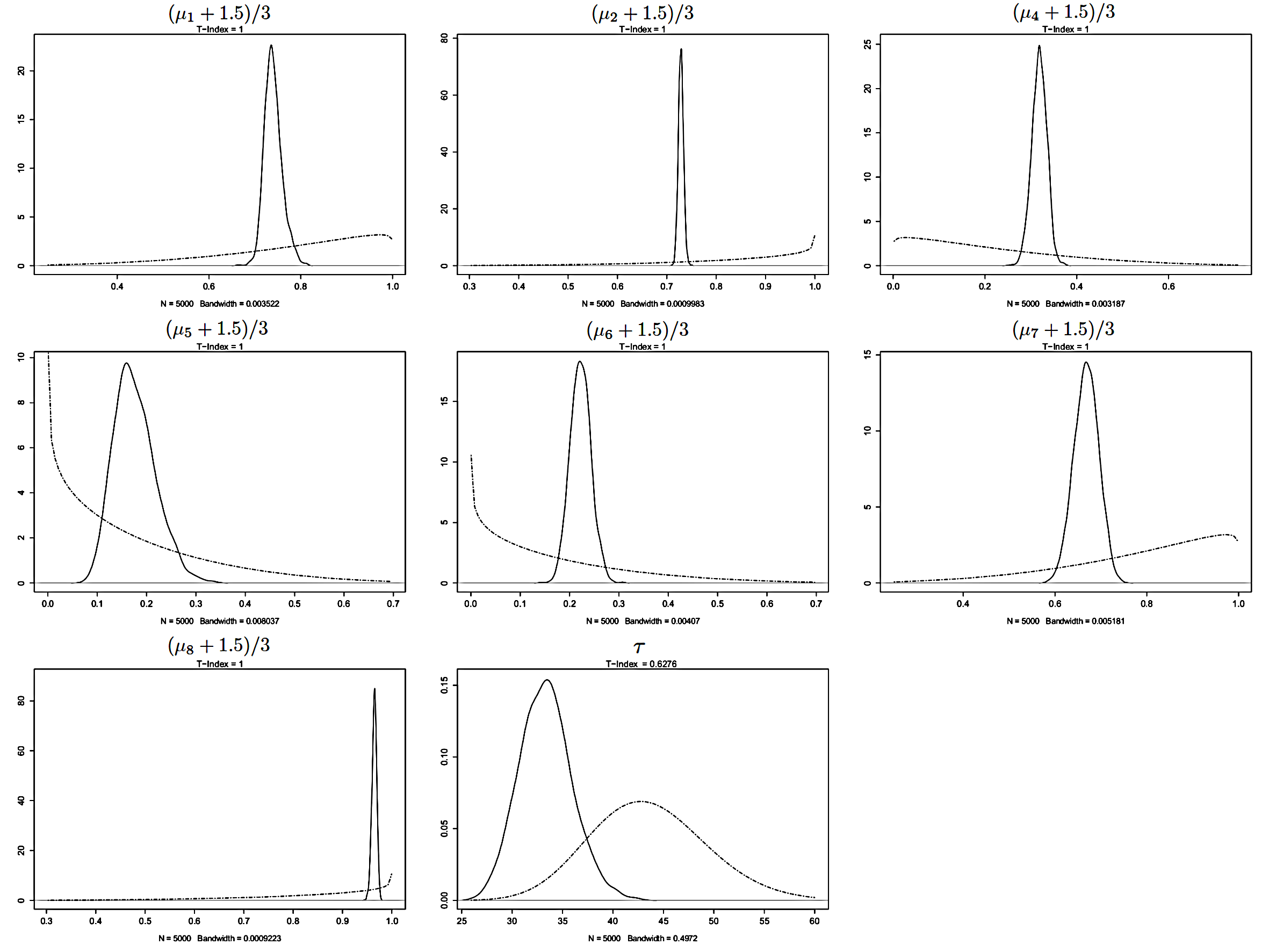}
\caption{Marginal probability density of parameters defining the conditional model of `Heart rate'.
Straight lines: posterior probability density. Dotted lines: prior probability density.
The D-statistic value is shown below the title.}
\label{fig:HR_mcmc}
\end{figure}

The probability distribution of `Heart rate' was defined by
applying the Beta regression model explained in Subsection
\ref{sub:betareg}:
The elicited prior distribution of parameters was:
\clearpage %%%%%%%%%%%%%%%%%%%%%%%%%%%%%%%%%%%%%%%%%%%%%%%%%%%%%%%%%
\begin{equation}
\begin{gathered}
\frac{\mu_0+1.5}{3} \sim \delta(0.5) \\
\frac{\mu_1+1.5}{3} \sim \text{Beta}(3.9187,1.0813)\\
\frac{\mu_2+1.5}{3} \sim \text{Beta}(4.1667,0.8333)\\
\frac{\mu_3+1.5}{3} \sim \delta(0.5)\\ 
\frac{\mu_4+1.5}{3} \sim \text{Beta}(1.0813,3.9187)\\
\frac{\mu_5+1.5}{3} \sim \text{Beta}(0.8333,4.1667)\\
\frac{\mu_6+1.5}{3} \sim \text{Beta}(0.8333,4.1667)\\
\frac{\mu_7+1.5}{3} \sim \text{Beta}(3.9187,1.0813)\\
\frac{\mu_8+1.5}{3} \sim \text{Beta}(4.1667,0.8333)\\
\tau \sim \text{Gamma}(89.4917,2.0304)
\end{gathered}
\end{equation}
where $\delta$ denotes a degenerated distribution (Dirac delta function).
In Figure \ref{fig:HR_mcmc}, marginal prior distribution and 
kernel density estimate of marginal posterior distribution
of model parameters are shown.
The summary of marginal prior and posterior distribution of
parameters is provided in Table \ref{tab:HR_mcmc}.

\begin{table}[!h]
\caption{Summary of marginal prior and posterior probability density of
parameters defining the conditional model of `Heart rate'.
`Std. dev.': standard deviation. `95\% QI': 95\% quantile interval.}
\label{tab:HR_mcmc}
\centering \footnotesize
\vspace{.3cm}
\begin{tabular}{crrrrrr}
 & Prior & Prior     & \multirow{2}{*}{Prior 95\% QI} & Posterior & Posterior & \multirow{2}{*}{Posterior 95\% QI}\\
 & mean  & std. dev. &                                & mean      & std. dev. &\\
\hline
$\frac{\mu_0+1.5}{3}$ & 0.5000 & 0.0000 & (0.5000, 0.5000) & 0.5000 & 0.0000 & (0.5000, 0.5000)\\
$\frac{\mu_1+1.5}{3}$ & 0.7837 & 0.0282 & (0.3770, 0.9913) & 0.7400 & 0.0194 & (0.7070, 0.7838)\\
$\frac{\mu_2+1.5}{3}$ & 0.8333 & 0.0231 & (0.4434, 0.9973) & 0.7287 & 0.0052 & (0.7187, 0.7391)\\
$\frac{\mu_3+1.5}{3}$ & 0.5000 & 0.0000 & (0.5000, 0.5000) & 0.5000 & 0.0000 & (0.5000, 0.5000)\\
$\frac{\mu_4+1.5}{3}$ & 0.2163 & 0.0282 & (0.0087, 0.6230) & 0.6816 & 0.0165 & (0.6511, 0.7149)\\
$\frac{\mu_5+1.5}{3}$ & 0.1667 & 0.0231 & (0.0027, 0.5566) & 0.8239 & 0.0421 & (0.7336, 0.8948)\\
$\frac{\mu_6+1.5}{3}$ & 0.1667 & 0.0231 & (0.0027, 0.5566) & 0.7779 & 0.0211 & (0.7350, 0.8184)\\
$\frac{\mu_7+1.5}{3}$ & 0.7837 & 0.0282 & (0.3770, 0.9913) & 0.6681 & 0.0268 & (0.6150, 0.7190)\\
$\frac{\mu_8+1.5}{3}$ & 0.8333 & 0.0231 & (0.4434, 0.9973) & 0.9644 & 0.0049 & (0.9543, 0.9733)\\
$\tau$ & 44.0759 & 4.6592 & (32.8782, 55.6731) & 33.3310 & 2.6107 & (28.4989, 38.7710)\\
\hline
\end{tabular}
\end{table}

%%%%%%%%%%%%%%%%%%%%%%%%%%%%%%%%%%%%%%%%%%%%%%%%%%%%%%%%%%%%%%%%

\section{Refinement}
\label{sec:refin}

After MCMC simulation, we implemented
the probabilistic network in $\mathsf{GeNIe}$ \citep{Druzdzel99}
by discretizing continuous variables into $5$ categories and by
computing CPTs at the posterior mean of parameters.
Afterwards, we performed two types of evaluation.
The first evaluation involved the Concordance Index
for all variables included in the
category $V_D$ (pathophysiology) with more than one quarter of
observed values and a percentage of non-neutral values greater
than $5\%$ (see Table \ref{tab:acutdis}).
The Concordance Index is defined as the proportion of patients
(judged as) affected by the disease with a predicted risk
greater than any patient not affected by the disease,
thus value $1$ indicates perfect discrimination of the judgement.
It was computed by considering data on clinical presentation only,
and by considering all the available patient data.
The second evaluation involved the inference performed by
the probabilistic network on six fictitious patient cases
elaborated by the second author.
These patient cases are described in Appendix \href{sec:appen1}.
If medical experts judged Concordance Index values and dia\-gnoses
satisfactory, the refinement process ended, otherwise medical
experts were invited to detect eventual inconsistencies with
medical causal knowledge in the qualitative part.
%on the basis of their clinical experience.
In this task, anchoring to parameters with a D-statistic
value less than $0.01$ was of great help for medical experts,
because an unsatisfactory diagnostic performance
of the probabilistic network often occurred together with a
strong divergence between the prior and the posterior distribution
of one or more parameters.
After that inconsistencies in the qualitative part
were detected, they were fixed and both elicitation and MCMC
simulation were repeated accordingly.

After refinement, our probabilistic network consisted of
$262$ variables, $574$ edges and $959$ parameters.
The frequency distribution of the D-statistic after
refinement is shown in Table \ref{tab:divergence}.

\begin{table}[!h]
\caption{Frequency distribution of the D-statistic after refinement.}
\label{tab:divergence}
\centering \small
\vspace{.3cm}
\begin{tabular}{cc}
Value of the D-statistic & Number of parameters\\ 
\hline
<0.01       & \phantom{0}55 \phantom{0}(5.73\%)\\
0.01-0.5    & \phantom{0}77 \phantom{0}(8.03\%)\\
0.5-0.925   &           151 (15.75\%)\\
0.925-0.975 &           192 (20.02\%)\\
>0.975      &           484 (51.47\%)\\
\hline
            & 959 \phantom{(00.00\%)}\\
\end{tabular}
\end{table}

Table \ref{tab:cindex} shows the estimation and $95\%$ confidence
interval (computed by bootstrapping) of the Concordance Index for
the selected acute diseases after refinement.
When considering data on clinical presentation only, Concordance Index
values are near or above $0.8$, suggesting a good diagnostic
performance \citep{Steyerberg10}.
When considering all the available patient data, Concordance Index values
are above $0.94$, confirming a substantial consistency between prior
knowledge encoded in the probabilistic network and data.

\begin{table}[!h]
\caption{Selected acute diseases.
$N_0$: number of patients without the disease.
$N_1$: number of patients with the disease.
'Total': number of patients on which the judgement was performed.}
\label{tab:acutdis}
\centering \small
\vspace{.3cm}
\begin{tabular}{lccc}
 & $N_0$ & $N_1$ & Total\\
\hline
Acute atrial fibrillation & 277 (90\%) & 31 (10\%) & 308\\
Congestive heart failure  & 599 (89\%) & 76 (11\%) & 675\\
Acute exacerbations of chronic bronchitis & 531 (86\%) & 85 (14\%)& 616\\
Angina without infarction & 242 (92\%) & 22\phantom{0} (8\%)  & 264\\
Myocardial infarction     & 629 (87\%) & 93 (13\%) & 722\\
Pneumonia                 & 424 (96\%) & 16\phantom{0} (4\%)  & 440\\
Pulmonary edema           & 617 (88\%) & 80 (12\%) & 697\\
Pulmonary embolism        & 654 (90\%) & 68 (10\%) & 722\\
Spontaneous pneumothorax  & 654 (97\%) & 23\phantom{0} (3\%)  & 677\\
\hline
\end{tabular}
\end{table}

\begin{table}[!h]
\caption{Concordance Index values for the selected of acute diseases
after refinement.
$95\%$ confidence intervals are computed by bootstrapping.}
\label{tab:cindex}
\centering \small
\vspace{.3cm}
\begin{tabular}{lcc}
 & Estimate & $95\%$ confidence interval\\
\hline
\multicolumn{3}{l}{\textbf{Considering clinical presentation only}}\\
\hspace{.15cm} Acute atrial fibrillation & 0.8076 & (0.6897, 0.9109)\\
\hspace{.15cm} Congestive heart failure  & 0.8770 & (0.8286, 0.9158)\\
\hspace{.15cm} Acute exacerbations of chronic bronchitis & 0.9324 & (0.8917, 0.9639)\\
\hspace{.15cm} Angina without infarction & 0.7905 & (0.6649, 0.8940)\\
\hspace{.15cm} Myocardial infarction     & 0.7524 & (0.6999, 0.8030)\\
\hspace{.15cm} Pneumonia                 & 0.8259 & (0.6784, 0.9436)\\
\hspace{.15cm} Pulmonary edema           & 0.8929 & (0.8537, 0.9238)\\
\hspace{.15cm} Pulmonary embolism        & 0.8207 & (0.7709, 0.8803)\\
\hspace{.15cm} Spontaneous pneumothorax  & 0.8565 & (0.7671, 0.9314)\\[3pt]
\multicolumn{3}{l}{\textbf{Considering all the available findings}}\\
\hspace{.15cm} Acute atrial fibrillation & 1.0000 & (1.0000, 1.0000)\\
\hspace{.15cm} Congestive heart failure  & 0.9609 & (0.9418, 0.9767)\\
\hspace{.15cm} Acute exacerbations of chronic bronchitis & 0.9084 & (0.8580, 0.9455)\\
\hspace{.15cm} Angina without infarction & 0.9027 & (0.8286, 0.9561)\\
\hspace{.15cm} Myocardial infarction     & 0.9834 & (0.9728, 0.9918)\\
\hspace{.15cm} Pneumonia                 & 0.9439 & (0.8976, 0.9787)\\
\hspace{.15cm} Pulmonary edema           & 0.9748 & (0.9625, 0.9873)\\
\hspace{.15cm} Pulmonary embolism        & 0.9742 & (0.9465, 0.9936)\\
\hspace{.15cm} Spontaneous pneumothorax  & 1.0000 & (1.0000, 1.0000)\\
\hline
\end{tabular}
\end{table}

Inference performed by the probabilistic network
on the six fictitious patient cases after refinement is shown in Appendix \href{sec:appen1}.

%%%%%%%%%%%%%%%%%%%%%%%%%%%%%%%%%%%%%%%%%%%%%%%%%%%%%%%%%%%%%%%%

\section{Discussion}
\label{sec:discuss}

In this paper, we described our experience in the development of
a probabilistic network for the diagnosis of acute cardiopulmonary
diseases.

In existing medical applications of probabilistic networks, the
qualitative part is often specified on the grounds of causal
knowledge documented in the specialised literature.
Several   advantages result from this causal formulation.
First, medical experts are able to understand
and discuss the information coded in the directed acyclic
graph (DAG), because it is expressed through notions provided
in standard medical training.
Second, it is widely recognized that the causal
formulation of DAGs often simplifies their specification (e.g., \cite{Koller09}, page 1009).
Third, causal DAGs are often characterized by a 
relatively small number of edges
while the same DAGs modified through arc reversal operations
typically contain more edges
(for example, see \cite{Buntine94}, Figure 16).
A comprehensive account of causal modelling is provided in \cite{Pearl09}.
In our work, we anchored the specification of the qualitative part
of the probabilistic network to the maximal constraint DAG, a
representation of equivalence classes of DAGs induced by relationships
among medical variables commonly accepted by physicians
(see the discussion in \cite{Luciani12}).

A second innovative aspect of our work is represented by the
use of both expert knowledge and patient data to estimate the
quantitative part of the proba\-bilistic network.
This approach is quite uncommon in the literature, where,
to our best knowledge, only one of the two
sources of information is typically exploited.
Remarkable applications include the Bayesian network developed
by \cite{Kline05}, and the expert systems DIAVAL \citep{Diez97}
and HEPAR II \citep{Wasyluk01}.
However, in the first work,a very small set of diseases is considered
and no prior information is exploited, as a consequence
the DAG results highly connected and estimates of parameters
are characterized by a large variance.
In the other two works, hundreds of variables are considered,
but expert knowledge is exploited only in the specification
of the qualitative part, while the quantitative part is estimated
from patient data only after continuous variables underwent to discretization.
In our proposal, discretization of continuous variables is avoided
by applying an original formal representation to the quantitative part,
consisting in a combination of the Beta regression and the
categorical logistic regression, reparameterized to help
physicians in performing quantitative assessments competently.
This way, the dimensionality of the quantitative part was reduced
to less than one thousand parameters for more than $260$ variables,
and medical experts were able to interpret parameters
with no additional training besides the standard medical one.
Furthermore, they were satisfied for the possibility
to express uncertainty in terms of the sample size
of pertaining clinical studies or, when unavailable, in
terms of the number of patient cases they experienced.
If this was the case, medical experts paid attention
to assess a number of patient cases substantially lower
than the one typically handled by clinical studies, in
fulfillment of principles of the Evidence-Based Medicine
(EBM) paradigm \citep{Guyatt92}.

Almost a quarter of variables included in the probabilistic
network was completely unobserved in our database,
while the mean percentage of missing va\-lues for those observed
was $34\%$, for a total of $53\%$ missing values among
all the va\-riables included in the probabilistic network.
Parameter estimation with unobserved variables is
typically challenging, as the related probability
distributions could not be consistently inferred from
collected data \citep{Settimi98}.
However, when the joint prior distribution of parameters
is informative like in our work, Bayesian estimation is
possible, and it is typically performed
by Markov Chain Monte Carlo (MCMC) simulation.
In this case, the marginal prior distribution of some parameters
remains unaltered after conditioning to collected data.
\citep{Gustafson15}.

The combination of two sources of information in MCMC
simulation, that is quantitative beliefs from experts and data,
helped medical experts in refining the probabilistic network.
During refinement, we often found that unsatisfactory
Concordance Index values for a selection of acute diseases
and/or unreasonable inference
on six fictitious patient cases occurred together with a
strong divergence between the prior and the posterior
distribution of one or more parameters (low value of the D-statistic).

After refinement, the probabilistic network consisted of
$262$ variables, $574$ edges and $959$ parameters.
In particular, it can be employed to perform medical
diagnosis on a total of 63 diseases (38 acute and 25 chronic)
on the basis of up to 167 patient findings.
The large set of diseases included in the probabilistic network 
highlights a further important feature of our work.
On one hand, this may be of help to prompt events not
considered by physicians, like rare diseases.
On the other hand, it may improve the diagnosis for patients
with an atypical presentation.

The refinement process ended after that the probabilistic network
provided satisfactory Concordance Index values for
a selection of acute diseases and after obtaining plausible inferences
on six fictitious patient cases.
Nevertheless, the empirical validation of its diagnostic performance
remains a mandatory objective before it may be used to support
decision making in a production environment.
At this purpose, future work will include %the design and analysis 
an evaluation of the probabilistic network based on data
perspectively collected.

%%%%%%%%%%%%%%%%%%%%%%%%%%%%%%%%%%%%%%%%%%%%%%%%%%%%%%%%%%%%%%%%%%%%%%%

\section*{Acknowledgments}

We thank Alessandro Squizzato, Andrea Rubboli, Leonardo Di Gennaro, Raffaele Landolfi,
Carlo De Luca, Fernando Porro, Marco Moia, Sophie Testa, Davide Imberti and Guido Bertolini
for their contribution.
This work was partially supported by the University of Florence, 
funding framework \textit{Progetto strategico di ricerca di base per l'anno 2015},
grant \textit{Disegno e analisi di studi sperimentali e osservazionali per le decisioni in ambito epidemiologico, socio-economico, ambientale e tecnologico}.
The authors declare that there are no conflicts of interest.
Financial disclosure: Sanofi-Aventis financially supported data collection.

%%%%%%%%%%%%%%%%%%%%%%%%%%%%%%%%%%%%%%%%%%%%%%%%%%%%%%%%%%%%%%%%%%%%%%%

\bibliographystyle{elsarticle-harv}
%\section*{\refname}
\bibliography{CardiopulmNet_bib}

\begin{thebibliography}{38}
\expandafter\ifx\csname natexlab\endcsname\relax\def\natexlab#1{#1}\fi
\expandafter\ifx\csname url\endcsname\relax
  \def\url#1{\texttt{#1}}\fi
\expandafter\ifx\csname urlprefix\endcsname\relax\def\urlprefix{URL }\fi

\bibitem[{Andreassen et~al.(1991)Andreassen, Hovorka, Benn, Olesen, and
  Carson}]{Andreassen91a}
Andreassen, S., Hovorka, R., Benn, J., Olesen, K.~G., Carson, E.~R., 1991. {A
  model-based approach to insulin adjustment}. In: Proceedings of the 3rd
  Conference on Artificial Intelligence in Medicine. Springer, Maastricht, NL,
  pp. 239--248.

\bibitem[{Buntine(1994)}]{Buntine94}
Buntine, W.~L., 1994. {Operations for learning with graphical models}. Journal
  of Artificial Intelligence Research 2, 159--225.

\bibitem[{Charitos et~al.(2009)Charitos, der Gaag, Visscher, Schurink, and
  Lucas}]{Charitos09}
Charitos, T., der Gaag, L. C.~V., Visscher, S., Schurink, K.~A., Lucas, P.~J.,
  2009. {A dynamic Bayesian network for diagnosing ventilator-associated
  pneumonia in ICU patients}. Expert Systems with Applications 36, 1249--1258.

\bibitem[{der Gaag et~al.(2002)der Gaag, Renooij, Witteman, Aleman, and
  Taal}]{Gaag02b}
der Gaag, L. C.~V., Renooij, S., Witteman, C. L.~M., Aleman, B. M.~P., Taal,
  B.~G., 2002. {Probabilities for a probabilistic network: a case-study in
  oesophageal carcinoma}. Artificial Intelligence in Medicine 25~(3), 123--148.

\bibitem[{D{\'i}ez et~al.(1997)D{\'i}ez, Mira, Iturralde, and
  Zubillaga}]{Diez97}
D{\'i}ez, F.~J., Mira, J., Iturralde, E., Zubillaga, S., 1997. {DIAVAL, a
  Bayesian expert system for echocardiography}. Artificial Intelligence in
  Medicine 10~(1), 59--73.

\bibitem[{Druzdzel(1999)}]{Druzdzel99}
Druzdzel, M.~J., 1999. {SMILE: Structural modeling, inference, and learning
  engine, and GeNIe: a development environment for graphical decision-theoretic
  models}. In: Proceedings of the National Conference on Artificial
  Intelligence. AAAI Press, Orlando, US-FL, pp. 902--903.

\bibitem[{Druzdzel and der Gaag(2000)}]{Druzdzel00a}
Druzdzel, M.~J., der Gaag, L. C.~V., 2000. {Building probabilistic networks:
  where do the numbers come from? A guide to the literature}. IEEE Transactions
  on Knowledge and Data Engineering 12~(4), 481--486.

\bibitem[{Ferrari and Cribari-Neto(2004)}]{Ferrari04}
Ferrari, S. L.~P., Cribari-Neto, F., 2004. {Beta regression for modelling rates
  and proportions}. Journal of Applied Statistics 31~(7), 799--815.

\bibitem[{Gal{\'a}n et~al.(2002)Gal{\'a}n, Aguado, D{\'i}ez, and
  Mira}]{Galan02}
Gal{\'a}n, S.~F., Aguado, F., D{\'i}ez, F.~J., Mira, J., 2002. {NasoNet:
  Modeling the spread of nasopharyngeal Cancer with networks of probabilistic
  events in discrete time}. Artificial Intelligence in Medicine 25~(3),
  27--264.

\bibitem[{Geweke(1992)}]{Geweke92}
Geweke, J., 1992. {Evaluating the accuracy of sampling-based approaches to
  calculating posterior moments}. In: Bernardo, J.~M., Berger, J.~O., Dawid,
  A.~P., Smith, A.~F. (Eds.), Bayesian Statistics 4. Clarendon Press, Oxford,
  UK.

\bibitem[{Gustafson(2015)}]{Gustafson15}
Gustafson, P., 2015. {Bayesian inference for partially identified models:
  exploring the limits of limited data}. Chapman and Hall/CRC, London, UK.

\bibitem[{Guyatt et~al.(1992)Guyatt, Cairns, and Churchill}]{Guyatt92}
Guyatt, G., Cairns, J., Churchill, D., 1992. {Evidence-based Medicine. A new
  approach to teaching the practice of Medicine}. Journal of the American
  Medical Association 268~(17), 2420--2425.

\bibitem[{Heidelberger and Welch(1983)}]{Heidelberger83}
Heidelberger, P., Welch, P.~D., 1983. {Simulation run length control in the
  presence of an initial transient}. Operations Research 31, 1109--1144.

\bibitem[{Irwin and Rippe(2011)}]{Irwin11}
Irwin, R.~S., Rippe, J.~M., 2011. {Irwin and Rippe's intensive care Medicine},
  7th Edition. Lippincott Williams \& Wilkins, Philadelphia, US-PA.

\bibitem[{Jacobs et~al.(2001)Jacobs, Oxley, and DeMott}]{Jacobs01}
Jacobs, D.~S., Oxley, D.~K., DeMott, W.~R., 2001. {Jacobs \& DeMott laboratory
  test handbook}, 5th Edition. Lexi-Comp, Cleveland, US-OH.

\bibitem[{Kahneman et~al.(1982)Kahneman, Slovic, and Tversky}]{Kahneman82}
Kahneman, D., Slovic, P., Tversky, A., 1982. {Judgement under uncertainty:
  heuristics and biases}. Cambridge University Press, Cambridge, UK.

\bibitem[{Kline et~al.(2005)Kline, Novobilski, Kabrhel, Richman, and
  Courtney}]{Kline05}
Kline, J.~A., Novobilski, A.~J., Kabrhel, C., Richman, P.~B., Courtney, D.~M.,
  2005. {Derivation and validation of a Bayesian network to predict pretest
  probability of venous thromboembolism}. Annals of Emergency Medicine 45~(3),
  282--290.

\bibitem[{Koller and Friedman(2009)}]{Koller09}
Koller, D., Friedman, N., 2009. {Probabilistic graphical models. Principles and
  techniques}. The MIT Press, Cambridge, US-MA.

\bibitem[{Korb and Nicholson(2010)}]{Korb10}
Korb, K.~B., Nicholson, A.~E., 2010. {Bayesian artificial intelligence}, 2nd
  Edition. Chapman \& Hall/CRC, Cambridge, UK.

\bibitem[{Lacave and D{\'i}ez(2003)}]{Lacave03}
Lacave, C., D{\'i}ez, F.~J., 2003. {Knowledge acquisition in PROSTANET: a
  Bayesian network for diagnosing prostate cancer}. Knowledge-Based Intelligent
  Information and Engineering Systems, Lecture Notes in Computer Science 2774,
  1345--1350.

\bibitem[{Leibovici et~al.(2007)Leibovici, Paul, Nielsen, Tacconelli, and
  Andreassen}]{Leibovici07}
Leibovici, L., Paul, M., Nielsen, A.~D., Tacconelli, E., Andreassen, S., 2007.
  {The TREAT project: decision support and prediction using causal
  probabilistic networks}. International Journal of Antimicrobial Agents
  30~(1), 93--102.

\bibitem[{Lucas et~al.(2004)Lucas, der Gaag, and Abu-Hanna}]{Lucas04}
Lucas, P.~J., der Gaag, L. C.~V., Abu-Hanna, A., 2004. {Bayesian networks in
  Biomedicine and health-care}. Artificial Intelligence in Medicine 30~(3),
  201--214.

\bibitem[{Luciani et~al.(2007)Luciani, Cavuto, Antiga, Miniati, Monti,
  Pistolesi, and Bertolini}]{Luciani07}
Luciani, D., Cavuto, S., Antiga, L., Miniati, M., Monti, S., Pistolesi, M.,
  Bertolini, G., 2007. {Bayes pulmonary embolism assisted diagnosis: a new
  expert system for clinical use}. Emergence Medicine Journal 24~(3), 157--164.

\bibitem[{Luciani and Stefanini(2012)}]{Luciani12}
Luciani, D., Stefanini, F.~M., 2012. {Automated interviews on clinical case
  reports to elicit directed acyclic graphs}. Artificial Intelligence in
  Medicine 55~(1), 1--11.

\bibitem[{McCullagh and Nelder(1989)}]{McCullagh89}
McCullagh, P., Nelder, J.~A., 1989. {Generalized linear models}, 2nd Edition.
  Chapman \& Hall/CRC, London, UK.

\bibitem[{Middleton et~al.(1991)Middleton, Shwe, Heckerman, Henrion, Horvitz,
  Lehmann, and Cooper}]{Middleton91}
Middleton, B., Shwe, M., Heckerman, D., Henrion, H., Horvitz, E., Lehmann, H.,
  Cooper, G., 1991. {Probabilistic diagnosis using a reformulation of the
  INTERNIST-1/QMR knowledge base: part II. Evaluation of diagnostic
  performance}. SIAM Journal on Computing 30, 256--267.

\bibitem[{Nathwani et~al.(1997)Nathwani, Clarke, Pike, and Azen}]{Nathwani97}
Nathwani, B.~N., Clarke, K., Pike, M.~C., Azen, S.~P., 1997. {Evaluation of an
  expert system on lymph node pathology}. Human Pathology 28~(9), 1097--1110.

\bibitem[{Pearl(2009)}]{Pearl09}
Pearl, J., 2009. {Causality: Models, reasoning, and inference}, 2nd Edition.
  Cambridge University Press, Cambridge, UK.

\bibitem[{Plummer(March 20-22, 2003)}]{Plummer03}
Plummer, M., March 20-22, 2003. {JAGS: A program for analysis of Bayesian
  graphical models using Gibbs sampling}. In: Proceedings of the 3rd
  International Workshop on Distributed Statistical Computing. Vienna, AT.

\bibitem[{Raftery and Lewis(1995)}]{Raftery95}
Raftery, A.~E., Lewis, S.~M., 1995. {The number of iterations, convergence
  diagnostics and generic Metropolis algorithms}. In: Gilks, W.~R.,
  Spiegelhalter, D.~J., Richardson, S. (Eds.), Practical Markov Chain Monte
  Carlo. Chapman and Hall, London, UK.

\bibitem[{Raghunathan(2004)}]{Raghunathan04}
Raghunathan, T.~E., 2004. {What do we do with missing data? Some options for
  analysis of incomplete data}. Annual Review of Public Health 25, 99--117.

\bibitem[{Settimi and Smith(1998)}]{Settimi98}
Settimi, R., Smith, J.~Q., 1998. {On the geometry of Bayesian graphical models
  with hidden variables}. In: Proceedings of the 14th Conference on Uncertainty
  in Artificial Intelligence. Morgan Kaufmann, Madison, US-WI, pp. 472--479.

\bibitem[{Squizzato et~al.(2011)Squizzato, Luciani, Rubboli, Gennaro, Landolfi,
  Luca, Porro, Moia, Testa, Imberti, and Bertolini}]{Squizzato11}
Squizzato, A., Luciani, D., Rubboli, A., Gennaro, L., Landolfi, R., Luca,
  C.~D., Porro, F., Moia, M., Testa, S., Imberti, D., Bertolini, G., 2011.
  {Differential diagnosis of pulmonary embolism in outpatients with
  non-specific cardiopulmonary symptoms}. Internal and Emergency Medicine,
  1--8.

\bibitem[{Steyerberg et~al.(2010)Steyerberg, Vickers, Cook, Gerds, Gonen,
  Obuchowski, Pencina, and Kattan}]{Steyerberg10}
Steyerberg, E.~W., Vickers, A.~J., Cook, N.~R., Gerds, T., Gonen, M.,
  Obuchowski, N., Pencina, M.~J., Kattan, M.~W., 2010. {Assessing the
  performance of prediction models: a framework for traditional and novel
  measures}. Epidemiology 21, 128--138.

\bibitem[{Suojanen et~al.(1999)Suojanen, Andreassen, and Olesen}]{Suojanen99}
Suojanen, M., Andreassen, S., Olesen, K.~G., 1999. {The EMG diagnosis: an
  interpretation based on partial information}. Medical Engineering and Physics
  21~(6-7), 517--523.

\bibitem[{Wasyluk et~al.(2001)Wasyluk, Onisko, and Druzdzel}]{Wasyluk01}
Wasyluk, H., Onisko, A., Druzdzel, M.~J., 2001. {Support of diagnosis of liver
  disorders based on a causal Bayesian network model}. Medical Science Monitor
  7~(1), 327--332.

\bibitem[{Winkler(1967)}]{Winkler67}
Winkler, R., 1967. {The assessment of prior distributions in Bayesian
  analysis}. Journal of the American Statistical Association 62~(319),
  776--780.

\bibitem[{Yuan and Druzdzel(2012)}]{Yuan12}
Yuan, C., Druzdzel, M.~J., 2012. {Importance sampling algorithms for Bayesian
  networks: principles and performance}. Mathematical and Computer Modelling
  43, 1189--1207.

\end{thebibliography}

\section*{Appendix 1. Fictitious patient cases}
\pdfbookmark{Appendix 1}{sec:appen1}

\paragraph{Case 1} %[S2]
A 73 years old man complained of mild fever and shortness of breath.
He had a history of chronic obstructive pulmonary disease and a
myocardial infarction ten years before.
On examination, he revealed bronchospasm and crackles in the lower
third of the lung.
Arterial pressure was 150/90 mmHg, heart rate 120 bpm.
The electrocardiogram showed a supraventricular arrithmya
apparently never occurred before.
On blood gas-analysis, oxygen saturation was 93\% after 4 L/min of
oxygen, and carbon dioxide arterial partial pressure was 50 mmHg.
Chest X-rays showed signs of pulmonary condensation.

\paragraph{Case 2} %[S2]
A 45 years old woman complained of acute chest pain and shortness of breath.
On examination, arterial pressure was 100/70 mmHg and heart rate was 90 bpm.
On blood gas-analysis, oxygen saturation was 94\% and carbon
dioxide arterial partial pressure was 34 mmHg.
Contraceptive pill apart, she did no take any drug, but she
had a smoker habit.
The D-dimer test was positive and the blood count was normal.
Chest X-rays revealed no anomalies.

\paragraph{Case 3} %[S1]
A 67 years old man, with chronic arterial hypertension and
smoking habit, suffered of oppressive chest pain, along with
sweetness and paleness.
His arterial pressure was 110/80 mmHg and heart rate was 90 bpm.
Laboratory tests showed a mild leucocytosis and abnormal Troponin I.
Chest X-rays was normal. 

\paragraph{Case 4} %[S2]
A 63 years old man with a long standing diabetes referred
to be recently collapsed.
On examination, glycemia was 145 mg/dl, heart rate was
110 bpm, arterial pressure was 100/60 mmHg and chest
X-rays was normal.
The electrocardiogram revealed a right bundle block and
sign of axis deviation.
The D-dimer test was positive and blood gas-analysis
showed an oxygen arterial partial pressure of 85 mmHg
and a carbon dioxide arterial partial pressure of 30 mmHg.

\paragraph{Case 5} %[S1]
A 50 years old man complained of fever since 3-4 days.
He had a mild cough and, more recently, he was affected
by mild confusion.
On examination, arterial pressure was 100/70 mmHg and heart
rate was 100 bpm.
On blood gas-analysis, oxygen saturation was 93\% and
hypocapnya (carbon dioxide arterial partial pressure was 30 mmHg).
Other laboratory tests showed normal hemoglobin, mild
leucocytosis and a mild increment of serum creatinine.
Chest X-rays showed an increment of the interstitial pulmonary net.

\paragraph{Case 6} %[S2]
A 85 years old woman, with cardiopathy, atrial fibrillation,
diabetes and chronic renal failure, complained of acute shortness
of breath.
On the laboratory tests, Brain Natriuretic Peptide was 200 pg/mL,
Troponin I was.5 mg/dL and serum creatinine was 1.9 mg/dL.
Heart rate was 98 bpm, oxygen saturation was 94\% after
administration of 2 L/min of oxygen.
On examination, crepitations emerged at the pulmonary basis,
confirmed by signs of pulmonary congestion at the chest X-rays.
On blood gas analysis, carbon dioxide arterial partial pressure was 70 mmHg.

\begin{table}[!h]
\caption{Inference performed by the probabilistic network on
the six fictitious patient cases after refinement.
Values represent probabilities.}
\label{tab:diagtest}
\centering \small
\vspace{.3cm}
\begin{tabular}{lcccccc}
 & Case 1 & Case 2 & Case 3 & Case 4 & Case 5 & Case 6 \\
\hline
Acute atrial fibrillation & 1.0000 & 0.0000 & 0.0000 & 0.0000 & 0.0000 & 1.0000\\
Congestive heart failure  & 0.4771 & 0.0019 & 0.0056 & 0.0004 & 0.6660 & 0.4864\\
Acute exacerbations of    & \multirow{2}{*}{0.9334} & \multirow{2}{*}{0.0259} & \multirow{2}{*}{0.0240} & \multirow{2}{*}{0.0093} & \multirow{2}{*}{0.3792} & \multirow{2}{*}{0.2654}\\
\hspace{.2cm} chronic bronchitis & & & & & &\\
Angina without infarction & 0.0098 & 0.3581 & 0.8951 & 0.0334 & 0.0064 & 0.1275\\
Myocardial infarction     & 0.0193 & 0.1281 & 0.8591 & 0.0103 & 0.0169 & 0.1826\\
Pneumonia                 & 0.6190 & 0.0045 & 0.0091 & 0.0048 & 0.2984 & 0.0643\\
Pulmonary edema           & 0.4966 & 0.0043 & 0.0058 & 0.0004 & 0.7379 & 0.5068\\
Pulmonary embolism        & 0.1130 & 0.4065 & 0.0088 & 0.2098 & 0.0340 & 0.0824\\
Spontaneous pneumothorax  & 0.0084 & 0.0000 & 0.0000 & 0.0000 & 0.0059 & 0.0029\\
\hline
\end{tabular}
\end{table}

~\\
\noindent
Table \ref{tab:diagtest} shows the diagnosis on the the six fictitious patient
cases performed by the probabilistic network after refinement.

In case 1, pneumonia appears the most likely hypothesis,
supported by the occurrence of fever, pulmonary crackles
on auscultation and, above all, consolidation in the chest film.
Absence of cough would represent an atypical finding, although the symptom
may have not been reported because deemed obvious in patients affected
by chronic obstructive pulmonary disease.
A cardiogenic pulmonary edema may also have occurred at this age, particularly
in the light of a past episode of ischemic heart disease.
Furthermore, the reported onset of supraventricular arrithmya may have
easily triggered congestive heart failure.
A chronic obstructive pulmonary disease exacerbation is likely to coexist
with both diagnostic hypotheses.

In case 2,
pulmonary embolism is the most likely diagnosis, due to the pre\-sence of a
risk factor (oestrogen assumption) combined with respiratory symptoms.
Nonetheless, the smoking habit and middle age of the patient also make an
acute coronary event and a subsequent myocardial infarction possible.

In case 3,
the presentation immediately reminds of cases with acute myocardial
infarction, as attested by the chest symptoms and the laboratory findings.
However, the model does not forget to remind of the existence of rarer
conditions, like pulmonary embolism.

In case 4,
after that electrocardiographic findings reveal a right axis deviation,
the diagnosis is correctly oriented towards pulmonary embolism.

In case 5,
the shortness of breath, together with the low oxygen saturation and the
increment of the interstitial pulmonary net, makes pulmonary edema a likely
hypothesis, possibly explained as the effect of an acute heart fai\-lure.
Nonetheless, other diagnostic hypotheses cannot be dismissed, parti\-cularly
pneumonia, which could justify the state of confusion, although similar
consequences on patient's awareness may be referred to the low oxygen saturation.

In case 6,
acute heart failure is by far the most likely explanation of worsening dyspnea
due to chronic heart disease.
As a secondary complicating disorder, pulmonary edema is suggested by chest
X-rays and the pulmonary crepitations on auscultation.
Underlying causes may encompass acute myocardial infarction, despite the absence
of chest pain, particularly because the patient is diabetic and troponin
levels are higher than normal.
However, elevated troponin levels may be explained by other hypotheses,
like a sustained tachycardia, although heart rate was not extremely
higher than normal at the time of visit.

%%%%%%%%%%%%%%%%%%%%%%%%%%%%%%%%%%%%%%%%%%%%%%%%%%%%%%%%%%%%%%%%%%%%%%%%%

~\\

\section*{Appendix 2. Parent set, typology and frequency distribution for each variable}
\pdfbookmark{Appendix 2}{sec:appen2}

Each variable included in the probabilistic network
is listed below alphabetically, %following the alphabetical order.
together with its typology (within round brackets), relative frequency
distribution (within square brackets) and parent set (after colons).
The symbol `-' indicates that a variable has no parent variables.\\

\hangindent=0.5cm \hangafter=1 \noindent
\textit{Abnormal ventilation trigger} (Pathophysiology, multi-valued) [absent: 0\%, present (hypo): 0\%, present (hyper): 0\%, missing: 100\%]: Pneumonia, Pulmonary edema, Pulmonary emphysema, Pulmonary venous thrombo-embolism. 
 
\hangindent=0.5cm \hangafter=1 \noindent 
\textit{Acute anemia} (Pathology, binary) [absent: 96.4\%, present: 1.8\%, missing: 1.8\%]: Hemorrhage. 
 
\hangindent=0.5cm \hangafter=1 \noindent 
\textit{Acute aortic valve failure} (Pathophysiology, binary) [absent: 15.51\%, present: 0.14\%, missing: 84.35\%]: Aortic dissection, Dilated cardiomyopathy, Endocarditis. 
 
\hangindent=0.5cm \hangafter=1 \noindent 
\textit{Acute atrial arrhythmia} (Pathology, binary) [absent: 38.37\%, present: 4.29\%, missing: 57.34\%]: Bradycardia/Tachycardia, Chronic atrial arrhythmia. 
 
\hangindent=0.5cm \hangafter=1 \noindent 
\textit{Acute cerebro-vascular disease} (Pathology, binary) [absent: 77.15\%, present: 2.22\%, missing: 20.64\%]: Arterial embolism, Arterial intra-vascular coagulation, Autonomic nervous system status, Chronic cerebro-vascular disease, Prophylaxis/anticoagulation. 
 
\hangindent=0.5cm \hangafter=1 \noindent 
\textit{Acute coronary event} (Pathology, binary) [absent: 26.45\%, present: 10.11\%, missing: 63.43\%]: Arterial intra-vascular coagulation, Chronic cardiac muscle  disease, Cocaine/amphetamines use, Right heart output. 
 
\hangindent=0.5cm \hangafter=1 \noindent 
\textit{Acute mitral valve failure} (Pathophysiology, binary) [absent: 17.59\%, present: 0.69\%, missing: 81.72\%]: Acute mitral valve prolapse, Acute myocardial infarction, Dilated cardiomyopathy, Endocarditis. 
 
\hangindent=0.5cm \hangafter=1 \noindent 
\textit{Acute mitral valve prolapse} (Pathophysiology, binary) [absent: 0\%, present: 0\%, missing: 100\%]: Acute myocardial infarction. 
 
\hangindent=0.5cm \hangafter=1 \noindent 
\textit{Acute myocardial infarction} (Pathology, multi-valued) [absent: 0\%, moderate: 0\%, severe: 0\%, missing: 100\%]: Acute coronary event, Myocarditis. 
 
\hangindent=0.5cm \hangafter=1 \noindent 
\textit{Acute pulmonary disease} (Pathophysiology, multi-valued) [absent: 73.55\%, initial: 4.02\%, advanced: 1.39\%, cardiac (asthma): 5.12\%, missing: 15.93\%]: Ashtma, Left heart pump, Lung cancer, Pneumonia, Pulmonary emphysema, Upper airways infection. 
 
\hangindent=0.5cm \hangafter=1 \noindent 
\textit{Acute respiratory distress syndrome} (Pathology, binary) [absent: 62.19\%, present: 0.55\%, missing: 37.26\%]: Lung cancer, Pancreatitis, Pneumonia, Pulmonary infarction, Sepsis. 
 
\hangindent=0.5cm \hangafter=1 \noindent 
\textit{Age (years old)} (Epidemiology, continuous) [<25: 0\%, 25-34: 0\%, 35-44: 0\%, 45-54: 0\%, 55-64: 0\%, 65-74: 0\%, 75-84: 0\%, 85-94: 0\%, 95-104: 0\%, >105: 0\%, missing: 100\%]: - 
 
\hangindent=0.5cm \hangafter=1 \noindent 
\textit{Air bronchogram} (Semiotics (other), binary) [absent: 15.79\%, present: 0.69\%, missing: 83.52\%]: Pulmonary consolidation. 
 
\hangindent=0.5cm \hangafter=1 \noindent 
\textit{Air trapping} (Semiotics (other), binary) [absent: 15.65\%, present: 0.69\%, missing: 83.66\%]: Pulmonary emphysema. 
 
\hangindent=0.5cm \hangafter=1 \noindent 
\textit{Alcoholism} (Aetiology, binary) [absent: 95.71\%, present: 3.19\%, missing: 1.11\%]: Age (years old). 
 
\hangindent=0.5cm \hangafter=1 \noindent 
\textit{Amylase} (Semiotics (other), binary) [normal: 56.37\%, augmented: 3.74\%, missing: 39.89\%]: Pancreatitis. 
 
\hangindent=0.5cm \hangafter=1 \noindent 
\textit{Anisosfigmia} (Semiotics (other), binary) [absent: 45.57\%, present: 0.28\%, missing: 54.16\%]: Aortic dissection. 
 
\hangindent=0.5cm \hangafter=1 \noindent 
\textit{Anti-inflammatory drugs recent intake} (Aetiology, binary) [no: 34.76\%, yes: 5.12\%, missing: 60.11\%]: Bacterial infection, Non-bacterial infection. 
 
\hangindent=0.5cm \hangafter=1 \noindent 
\textit{Antiphospholipids} (Semiotics (other), binary) [absent: 2.22\%, present: 0.28\%, missing: 97.51\%]: Thrombophilia. 
 
\hangindent=0.5cm \hangafter=1 \noindent 
\textit{Antithrombin III} (Semiotics (other), binary) [normal: 6.93\%, deficit: 0.69\%, missing: 92.38\%]: Thrombophilia. 
 
\hangindent=0.5cm \hangafter=1 \noindent 
\textit{Anxiety/agitation} (Pathophysiology, multi-valued) [absent: 0\%, moderate: 0\%, severe: 0\%, missing: 100\%]: Chest pain, Hypoglycemia. 
 
\hangindent=0.5cm \hangafter=1 \noindent 
\textit{Anxiety/agitation (according to clinical judgement)} (Semiotics (other), binary) [absent: 92.24\%, present: 6.65\%, missing: 1.11\%]: Anxiety/agitation. 
 
\hangindent=0.5cm \hangafter=1 \noindent 
\textit{Aortic aneurysm} (Epidemiology, binary) [absent: 85.18\%, present: 5.68\%, missing: 9.14\%]: Age (years old), Chronic arterial hypertension, Gender. 
 
\hangindent=0.5cm \hangafter=1 \noindent 
\textit{Aortic dissection} (Pathology, binary) [absent: 0\%, present: 0\%, missing: 100\%]: Age (years old), Aortic aneurysm, Gender. 
 
\hangindent=0.5cm \hangafter=1 \noindent 
\textit{Aortic intramural hematoma} (Semiotics (other), binary) [absent: 49.17\%, present: 0\%, missing: 50.83\%]: Aortic dissection. 
 
\hangindent=0.5cm \hangafter=1 \noindent 
\textit{Aortic stenosis} (Semiotics (other), binary) [absent: 0\%, present: 0\%, missing: 100\%]: - 
 
\hangindent=0.5cm \hangafter=1 \noindent 
\textit{Aortic valve failure} (Semiotics (other), binary) [absent: 28.95\%, present: 11.91\%, missing: 59.14\%]: Acute aortic valve failure, Chronic aortic valve failure. 
 
\hangindent=0.5cm \hangafter=1 \noindent 
\textit{Arterial embolism} (Pathogenesis, binary) [absent: 0\%, present: 0\%, missing: 100\%]: Arterial intra-vascular coagulation, Patent foramen ovale, Pulmonary venous thrombo-embolism. 
 
\hangindent=0.5cm \hangafter=1 \noindent 
\textit{Arterial intra-vascular coagulation} (Pathogenesis, multi-valued) [absent: 0\%, coronaric: 1.52\%, cerebral: 0.42\%, left (heart): 0\%, mixed: 0\%, missing: 98.06\%]: Aortic dissection, Chronic atrial arrhythmia, Chronic cardiac muscle  disease, Intra-vascular coagulation, Pancreatitis. 
 
\hangindent=0.5cm \hangafter=1 \noindent 
\textit{Arterial vascular resistance} (Pathophysiology, continuous) [lp-range: 0\%, n-range: 0\%, hp-range: 0\%, missing: 100\%]: Autonomic nervous system status, Chronic arterial hypertension, Left heart pump. 
 
\hangindent=0.5cm \hangafter=1 \noindent 
\textit{Ascending aorta intimal flap} (Semiotics (other), binary) [absent: 49.31\%, present: 0\%, missing: 50.69\%]: Aortic dissection. 
 
\hangindent=0.5cm \hangafter=1 \noindent 
\textit{Ashtma} (Pathology, binary) [absent: 78.53\%, present: 1.8\%, missing: 19.67\%]: Age (years old), Gender. 
 
\hangindent=0.5cm \hangafter=1 \noindent 
\textit{Atelactasis} (Semiotics (other), binary) [absent: 88.64\%, present: 6.23\%, missing: 5.12\%]: Acute respiratory distress syndrome, Lung cancer, Pulmonary consolidation, Spontaneous pneumothorax. 
 
\hangindent=0.5cm \hangafter=1 \noindent 
\textit{Atrial arrhythmia} (Semiotics (other), binary) [absent: 72.44\%, present: 26.45\%, missing: 1.11\%]: Acute atrial arrhythmia, Chronic atrial arrhythmia. 
 
\hangindent=0.5cm \hangafter=1 \noindent 
\textit{Augmented lactates (according to clinical judgment)} (Semiotics (other), binary) [no: 70.5\%, yes: 4.43\%, missing: 25.07\%]: Lactates (mmol/l). 
 
\hangindent=0.5cm \hangafter=1 \noindent 
\textit{Autonomic nervous system status} (Pathophysiology, multi-valued) [normal: 5.26\%, adrenergic (1): 0\%, adrenergic (2): 0\%, hypertensive (crisis): 1.25\%, cholinergic (1): 0\%, cholinergic (2): 0\%, missing: 93.49\%]: Acute coronary event, Acute myocardial infarction, Acute pulmonary disease, Anxiety/agitation, Chronic cardiac muscle  disease, Heart drive, Left heart pump, Right heart output, Sepsis. 
 
\hangindent=0.5cm \hangafter=1 \noindent 
\textit{Bacterial infection} (Pathophysiology, binary) [absent: 0\%, present: 0\%, missing: 100\%]: Cholecystitis, Endocarditis, Immunocompromission, Myocarditis, Non-infarctual pericarditis, Pancreatitis, Peritonitis, Pneumonia, Pulmonary infarction, Sepsis. 
 
\hangindent=0.5cm \hangafter=1 \noindent 
\textit{Bias of perfusion scintigraphy} (Pathophysiology, binary) [absent: 0\%, present: 0\%, missing: 100\%]: Atelactasis, Pleural effusion, Pulmonary consolidation, Pulmonary opacity. 
 
\hangindent=0.5cm \hangafter=1 \noindent 
\textit{Biliary colic} (Semiotics (other), binary) [absent: 96.12\%, present: 1.8\%, missing: 2.08\%]: Cholelithiasis. 
 
\hangindent=0.5cm \hangafter=1 \noindent 
\textit{Blood pressure (mmHg)} (Semiotics (other), continuous) [lp-range: 7.89\%, n-range: 43.49\%, hp-range: 47.23\%, missing: 1.39\%]: Arterial vascular resistance, Left cardiac output. 
 
\hangindent=0.5cm \hangafter=1 \noindent 
\textit{Bradycardia/Tachycardia} (Pathology, multi-valued) [absent: 64.13\%, bradycardia: 1.66\%, tachycardia (1): 3.32\%, tachycardia (2): 3.6\%, missing: 27.29\%]: Dehydration, Heart drive. 
 
\hangindent=0.5cm \hangafter=1 \noindent 
\textit{Brain natriuretic peptide} (Semiotics (other), binary) [normal: 0.28\%, augmented: 1.52\%, missing: 98.2\%]: Left heart pump. 
 
\hangindent=0.5cm \hangafter=1 \noindent 
\textit{Bronchial diameter} (Semiotics (other), binary) [normal: 14.68\%, reduced: 0.42\%, missing: 84.9\%]: Pulmonary emphysema. 
 
\hangindent=0.5cm \hangafter=1 \noindent 
\textit{Bronchial walls} (Semiotics (other), binary) [normal: 13.99\%, tickened: 1.39\%, missing: 84.63\%]: Pulmonary emphysema. 
 
\hangindent=0.5cm \hangafter=1 \noindent 
\textit{Bronchiectasis} (Semiotics (other), binary) [absent: 78.12\%, present: 1.8\%, missing: 20.08\%]: Pulmonary emphysema. 
 
\hangindent=0.5cm \hangafter=1 \noindent 
\textit{Bronchospasm/reduced vescicolar murmur} (Semiotics (other), multi-valued) [normal: 76.04\%, rhonchi (or wheezing): 19.25\%, silence: 2.08\%, missing: 2.63\%]: Acute pulmonary disease, Pulmonary emphysema, Pulmonary venous thrombo-embolism, Spontaneous pneumothorax, Upper airways infection. 
 
\hangindent=0.5cm \hangafter=1 \noindent 
\textit{Cardiac axis right deviation (S1-Q3-T3)} (Semiotics (other), binary) [absent: 90.58\%, present: 5.12\%, missing: 4.29\%]: ECG right heart findings. 
 
\hangindent=0.5cm \hangafter=1 \noindent 
\textit{Cardiac tamponade} (Pathophysiology, multi-valued) [absent: 0\%, moderate: 0\%, severe: 0\%, missing: 100\%]: Hemopericardium, Pericarditis. 
 
\hangindent=0.5cm \hangafter=1 \noindent 
\textit{Cardiomegaly} (Semiotics (other), binary) [absent: 55.96\%, present: 31.99\%, missing: 12.05\%]: Chronic cardiac muscle  disease, Left heart pump, Pericardial effusion. 
 
\hangindent=0.5cm \hangafter=1 \noindent 
\textit{Carotid sinus massage test} (Semiotics (other), binary) [negative: 0.28\%, positive: 0.69\%, missing: 99.03\%]: Sick sinus syndrome. 
 
\hangindent=0.5cm \hangafter=1 \noindent 
\textit{Cavitation/Colliquation} (Semiotics (other), binary) [absent: 15.65\%, present: 0.14\%, missing: 84.21\%]: Pneumonia. 
 
\hangindent=0.5cm \hangafter=1 \noindent 
\textit{Central cyanosis} (Semiotics (other), binary) [absent: 34.35\%, present: 0.55\%, missing: 65.1\%]: Oxygen saturation (percentage). 
 
\hangindent=0.5cm \hangafter=1 \noindent 
\textit{Central line} (Aetiology, binary) [no: 97.78\%, yes: 2.22\%, missing: 0\%]: - 
 
\hangindent=0.5cm \hangafter=1 \noindent 
\textit{Central mass (thoracic)} (Semiotics (other), binary) [absent: 88.09\%, present: 3.05\%, missing: 8.86\%]: Lung cancer. 
 
\hangindent=0.5cm \hangafter=1 \noindent 
\textit{Cerebral hypoxia} (Pathophysiology, multi-valued) [absent: 0\%, recently occurred: 0\%, present: 0\%, missing: 100\%]: Acute anemia, Acute atrial arrhythmia, Acute cerebro-vascular disease, Dehydration, Left cardiac output, Obstruction of the systemic circulation, Sick sinus syndrome, Temporary suspension of heart drive, Vasovagal syncope. 
 
\hangindent=0.5cm \hangafter=1 \noindent 
\textit{Cerebral mass} (Pathophysiology, binary) [absent: 70.91\%, present: 0.69\%, missing: 28.39\%]: Acute cerebro-vascular disease, Neoplastic disease (generic). 
 
\hangindent=0.5cm \hangafter=1 \noindent 
\textit{Chest pain} (Semiotics (other), binary) [absent: 39.47\%, present: 47.09\%, missing: 13.43\%]: Chest pain type. 
 
\hangindent=0.5cm \hangafter=1 \noindent 
\textit{Chest pain (gastro-oesophageal origin)} (Pathophysiology, binary) [absent: 93.91\%, present: 0.28\%, missing: 5.82\%]: Gastro-oesophageal reflux, Hiatal hernia, Mallory-Weiss syndrome. 
 
\hangindent=0.5cm \hangafter=1 \noindent 
\textit{Chest pain (parietal origin)} (Pathophysiology, binary) [absent: 0\%, present: 0\%, missing: 100\%]: Costochondritis, Herpes Zooster, Lower limbs fractures, Lung cancer, Pneumonia, Rib fracture, Spontaneous pneumothorax. 
 
\hangindent=0.5cm \hangafter=1 \noindent 
\textit{Chest pain (pleuritic origin)} (Pathophysiology, binary) [absent: 0\%, present: 0\%, missing: 100\%]: Costochondritis, Lower limbs fractures, Pleurisy, Pneumonia, Spontaneous pneumothorax. 
 
\hangindent=0.5cm \hangafter=1 \noindent 
\textit{Chest pain (retro-sternal origin)} (Pathophysiology, binary) [absent: 0\%, present: 0\%, missing: 100\%]: Aortic dissection, Chest pain (gastro-oesophageal origin), Chronic mitral valve prolapse, Dilatated pulmonary artery disease, Obstructive cardiomyopathy, Pericarditis, Pneumonia. 
 
\hangindent=0.5cm \hangafter=1 \noindent 
\textit{Chest pain (stabbing origin)} (Pathophysiology, binary) [absent: 0\%, present: 0\%, missing: 100\%]: Aortic dissection, Spontaneous pneumothorax. 
 
\hangindent=0.5cm \hangafter=1 \noindent 
\textit{Chest pain (upper-abdominal origin)} (Pathophysiology, binary) [absent: 0\%, present: 0\%, missing: 100\%]: Biliary colic, Chest pain (gastro-oesophageal origin), Pancreatitis, Peptic ulcer, Peritonitis. 
 
\hangindent=0.5cm \hangafter=1 \noindent 
\textit{Chest pain type} (Semiotics (other), multi-valued) [absent: 39.75\%, stabbing: 1.8\%, retro (sternal): 21.61\%, pleuritic: 7.76\%, upper (abdominal): 4.85\%, parietal: 4.71\%, missing: 19.53\%]: Acute coronary event, Chest pain (parietal origin), Chest pain (pleuritic origin), Chest pain (retro-sternal origin), Chest pain (stabbing origin), Chest pain (upper-abdominal origin). 
 
\hangindent=0.5cm \hangafter=1 \noindent 
\textit{Cholecystitis} (Pathology, binary) [absent: 0\%, present: 0\%, missing: 100\%]: Cholelithiasis. 
 
\hangindent=0.5cm \hangafter=1 \noindent 
\textit{Cholelithiasis} (Epidemiology, binary) [absent: 69.94\%, present: 4.85\%, missing: 25.21\%]: Age (years old), Gender, Obesity (Body Mass Index>=30). 
 
\hangindent=0.5cm \hangafter=1 \noindent 
\textit{Chronic anemia} (Epidemiology, binary) [absent: 0\%, present: 0\%, missing: 100\%]: Age (years old), Fertility. 
 
\hangindent=0.5cm \hangafter=1 \noindent 
\textit{Chronic aortic valve failure} (Epidemiology, multi-valued) [absent: 0\%, initial: 0\%, advanced: 0\%, missing: 100\%]: Age (years old), Aortic aneurysm, Dilated cardiomyopathy. 
 
\hangindent=0.5cm \hangafter=1 \noindent 
\textit{Chronic arterial hypertension} (Epidemiology, binary) [absent: 0\%, present: 0\%, missing: 100\%]: Age (years old), Gender, Obesity (Body Mass Index>=30), Smoker. 
 
\hangindent=0.5cm \hangafter=1 \noindent 
\textit{Chronic atrial arrhythmia} (Epidemiology, binary) [absent: 0\%, present: 0\%, missing: 100\%]: Age (years old), Chronic cardiac muscle  disease, Pulmonary emphysema. 
 
\hangindent=0.5cm \hangafter=1 \noindent 
\textit{Chronic cardiac muscle  disease} (Epidemiology, multi-valued) [absent: 0\%, initial: 0\%, advanced: 0\%, missing: 100\%]: Age (years old), Chronic arterial hypertension, Cor pulmonale, Dilated cardiomyopathy, Gender, Left ventricular hypertrophy. 
 
\hangindent=0.5cm \hangafter=1 \noindent 
\textit{Chronic cerebro-vascular disease} (Epidemiology, binary) [absent: 35.6\%, present: 0.69\%, missing: 63.71\%]: Age (years old), Chronic arterial hypertension. 
 
\hangindent=0.5cm \hangafter=1 \noindent 
\textit{Chronic interstitial lung disease} (Epidemiology, binary) [absent: 37.26\%, present: 0.28\%, missing: 62.47\%]: Age (years old). 
 
\hangindent=0.5cm \hangafter=1 \noindent 
\textit{Chronic metabolic alkalosis} (Epidemiology, binary) [absent: 0\%, present: 0\%, missing: 100\%]: Age (years old). 
 
\hangindent=0.5cm \hangafter=1 \noindent 
\textit{Chronic mitral valve failure} (Epidemiology, multi-valued) [absent: 0\%, initial: 0\%, advanced: 0\%, missing: 100\%]: Age (years old), Aortic stenosis, Chronic aortic valve failure, Chronic mitral valve prolapse, Dilated cardiomyopathy. 
 
\hangindent=0.5cm \hangafter=1 \noindent 
\textit{Chronic mitral valve prolapse} (Epidemiology, binary) [absent: 0\%, present: 0\%, missing: 100\%]: - 
 
\hangindent=0.5cm \hangafter=1 \noindent 
\textit{Chronic obstructive pulmonary disease} (Semiotics (other), multi-valued) [absent: 75.62\%, initial: 16.07\%, advanced: 0.97\%, missing: 7.34\%]: Pulmonary emphysema. 
 
\hangindent=0.5cm \hangafter=1 \noindent 
\textit{Chronic venous insufficiency} (Aetiology, binary) [absent: 63.43\%, present: 9.56\%, missing: 27.01\%]: Age (years old), Gender. 
 
\hangindent=0.5cm \hangafter=1 \noindent 
\textit{CK-MB} (Semiotics (other), binary) [normal: 29.5\%, augmented: 10.25\%, missing: 60.25\%]: Acute myocardial infarction. 
 
\hangindent=0.5cm \hangafter=1 \noindent 
\textit{Cocaine/amphetamines use} (Aetiology, binary) [no: 97.92\%, yes: 0.69\%, missing: 1.39\%]: - 
 
\hangindent=0.5cm \hangafter=1 \noindent 
\textit{Compression stockings} (Epidemiology, binary) [no: 98.75\%, yes: 1.25\%, missing: 0\%]: Surgery. 
 
\hangindent=0.5cm \hangafter=1 \noindent 
\textit{Confusion} (Semiotics (other), binary) [absent: 92.66\%, present: 7.34\%, missing: 0\%]: Glasgow Coma Scale, Previous transient seizure. 
 
\hangindent=0.5cm \hangafter=1 \noindent 
\textit{Cor pulmonale} (Epidemiology, binary) [absent: 0\%, present: 0\%, missing: 100\%]: Pulmonary hypertension. 
 
\hangindent=0.5cm \hangafter=1 \noindent 
\textit{Costochondritis} (Pathology, binary) [absent: 75.62\%, present: 0.14\%, missing: 24.24\%]: - 
 
\hangindent=0.5cm \hangafter=1 \noindent 
\textit{Cough} (Semiotics (other), multi-valued) [absent: 84.63\%, dry: 6.37\%, productive: 9\%, missing: 0\%]: Acute pulmonary disease, Pleurisy, Pneumonia, Pulmonary edema, Pulmonary emphysema, Upper airways infection. 
 
\hangindent=0.5cm \hangafter=1 \noindent 
\textit{Cystic areas /Bullae} (Semiotics (other), binary) [absent: 13.99\%, present: 1.52\%, missing: 84.49\%]: Pulmonary emphysema. 
 
\hangindent=0.5cm \hangafter=1 \noindent 
\textit{D-dimer test} (Semiotics (other), binary) [negative: 17.87\%, positive: 20.78\%, missing: 61.36\%]: Age (years old), Fibrinolysis, Pregnancy, Prophylaxis/anti- coagulation. 
 
\hangindent=0.5cm \hangafter=1 \noindent 
\textit{Dehydration} (Pathophysiology, continuous) [n-range: 0\%, hp-range: 0\%, missing: 100\%]: Pancreatitis, Sepsis. 
 
\hangindent=0.5cm \hangafter=1 \noindent 
\textit{Dilatated pulmonary artery disease} (Pathophysiology, multi-valued) [absent: 0\%, acute: 0\%, chronic: 0\%, missing: 100\%]: Pulmonary hypertension, Pulmonary venous thrombo-embolism. 
 
\hangindent=0.5cm \hangafter=1 \noindent 
\textit{Dilated ascending aorta} (Semiotics (other), binary) [absent: 85.18\%, present: 5.68\%, missing: 9.14\%]: Aortic aneurysm. 
 
\hangindent=0.5cm \hangafter=1 \noindent 
\textit{Dilated cardiomyopathy} (Pathology, binary) [absent: 29.22\%, present: 1.11\%, missing: 69.67\%]: Age (years old), Gender. 
 
\hangindent=0.5cm \hangafter=1 \noindent 
\textit{Dilated left ventricle} (Semiotics (other), binary) [absent: 31.44\%, present: 8.03\%, missing: 60.53\%]: Dilated cardiomyopathy, Left ventricular pre-load, Myocarditis. 
 
\hangindent=0.5cm \hangafter=1 \noindent 
\textit{Dilated pulmonary artery} (Semiotics (other), binary) [negative: 41.14\%, positive: 1.39\%, missing: 57.48\%]: Dilatated pulmonary artery disease. 
 
\hangindent=0.5cm \hangafter=1 \noindent 
\textit{Dilated right ventricle} (Semiotics (other), binary) [absent: 37.53\%, present: 5.26\%, missing: 57.2\%]: Cardiac tamponade, Dilated cardiomyopathy, Right heart pre-load. 
 
\hangindent=0.5cm \hangafter=1 \noindent 
\textit{Dyspepsia} (Pathophysiology, multi-valued) [absent: 0\%, moderate: 0\%, severe: 0\%, missing: 100\%]: Acute myocardial infarction, Biliary colic, Gastro-oesophageal reflux, Peptic ulcer. 
 
\hangindent=0.5cm \hangafter=1 \noindent 
\textit{Dyspnea} (Semiotics (future), binary) [absent: 29.5\%, present: 49.58\%, missing: 20.91\%]: Acute anemia, Anxiety/agitation, Lactates (mmol/l), Lung perfusion, Pulmonary shunt. 
 
\hangindent=0.5cm \hangafter=1 \noindent 
\textit{ECG right heart findings} (Pathophysiology, binary) [absent: 75.48\%, present: 12.05\%, missing: 12.47\%]: Acute coronary event, Right heart pre-load. 
 
\hangindent=0.5cm \hangafter=1 \noindent 
\textit{Elevated hemidiaphragm} (Semiotics (other), binary) [absent: 89.2\%, present: 5.68\%, missing: 5.12\%]: Atelactasis, Pulmonary infarction. 
 
\hangindent=0.5cm \hangafter=1 \noindent 
\textit{Endocardial vegetations} (Semiotics (other), binary) [absent: 43.63\%, present: 0.28\%, missing: 56.09\%]: Endocarditis. 
 
\hangindent=0.5cm \hangafter=1 \noindent 
\textit{Endocarditis} (Pathology, binary) [absent: 18.42\%, present: 0.28\%, missing: 81.3\%]: - 
 
\hangindent=0.5cm \hangafter=1 \noindent 
\textit{Endoluminal thrombus} (Semiotics (other), multi-valued) [normal: 8.17\%, subsegmental (arteries): 3.88\%, segmental (arteries): 4.43\%, missing: 83.52\%]: Pulmonary venous thrombo-embolism. 
 
\hangindent=0.5cm \hangafter=1 \noindent 
\textit{Extrasystoles} (Semiotics (other), binary) [absent: 0\%, present: 0\%, missing: 100\%]: Chronic mitral valve prolapse, Ventricular arrhythmia. 
 
\hangindent=0.5cm \hangafter=1 \noindent 
\textit{Extrogens use} (Aetiology, binary) [no: 98.48\%, yes: 1.52\%, missing: 0\%]: Fertility. 
 
\hangindent=0.5cm \hangafter=1 \noindent 
\textit{Factor II G20210A} (Semiotics (other), binary) [absent: 0\%, present: 0\%, missing: 100\%]: Thrombophilia. 
 
\hangindent=0.5cm \hangafter=1 \noindent 
\textit{Factor VIII} (Semiotics (other), binary) [normal: 0\%, augmented: 0\%, missing: 100\%]: Thrombophilia. 
 
\hangindent=0.5cm \hangafter=1 \noindent 
\textit{Fall-down} (Semiotics (other), binary) [absent: 84.49\%, present: 15.51\%, missing: 0\%]: Acute cerebro-vascular disease, Syncope. 
 
\hangindent=0.5cm \hangafter=1 \noindent 
\textit{Fertility} (Aetiology, binary) [no: 0\%, yes: 0\%, missing: 100\%]: Age (years old), Gender. 
 
\hangindent=0.5cm \hangafter=1 \noindent 
\textit{Fever} (Semiotics (other), binary) [absent: 39.47\%, present: 11.22\%, missing: 49.31\%]: Anti-inflammatory drugs recent intake, Bacterial infection, Non-bacterial infection. 
 
\hangindent=0.5cm \hangafter=1 \noindent 
\textit{Fibrinolysis} (Pathogenesis, binary) [absent: 0\%, present: 0\%, missing: 100\%]: Arterial intra-vascular coagulation, Venous intra-vascular coagulation. 
 
\hangindent=0.5cm \hangafter=1 \noindent 
\textit{Focal neurological signs} (Semiotics (other), binary) [absent: 96.4\%, present: 3.6\%, missing: 0\%]: Acute cerebro-vascular disease, Chronic cerebro-vascular disease. 
 
\hangindent=0.5cm \hangafter=1 \noindent 
\textit{Focal neurological signs} (Semiotics (other), binary) [absent: 96.4\%, present: 3.6\%, missing: 0\%]: Acute cerebro-vascular disease, Chronic cerebro-vascular disease. 
 
\hangindent=0.5cm \hangafter=1 \noindent 
\textit{FT3 (pg/ml)} (Semiotics (other), continuous) [lp-range: 8.73\%, n-range: 10.94\%, hp-range: 0.55\%, missing: 79.78\%]: Thyroid hormones. 
 
\hangindent=0.5cm \hangafter=1 \noindent 
\textit{FT4 (ng/ml)} (Semiotics (other), continuous) [lp-range: 0.97\%, n-range: 20.78\%, hp-range: 1.39\%, missing: 76.87\%]: Thyroid hormones. 
 
\hangindent=0.5cm \hangafter=1 \noindent 
\textit{Gastro-oesophageal reflux} (Pathology, binary) [absent: 0\%, present: 0\%, missing: 100\%]: Age (years old), Hiatal hernia. 
 
\hangindent=0.5cm \hangafter=1 \noindent 
\textit{Gender} (Epidemiology, binary) [male: 52.22\%, female: 47.37\%, missing: 0.42\%]: - 
 
\hangindent=0.5cm \hangafter=1 \noindent 
\textit{Generalized epileptic seizure} (Semiotics (other), binary) [absent: 97.78\%, present: 1.66\%, missing: 0.55\%]: Previous transient seizure. 
 
\hangindent=0.5cm \hangafter=1 \noindent 
\textit{Glasgow Coma Scale} (Semiotics (other), multi-valued) [absent: 93.49\%, from (12 to 14): 4.16\%, from (9 to 11): 0.55\%, less (than 9): 1.52\%, missing: 0.28\%]: Cerebral hypoxia, Cerebral mass, Hypoglycemia, Oxygen saturation (percentage). 
 
\hangindent=0.5cm \hangafter=1 \noindent 
\textit{Ground Glass} (Semiotics (other), binary) [absent: 13.3\%, present: 0.55\%, missing: 86.15\%]: Pulmonary edema, Pulmonary emphysema. 
 
\hangindent=0.5cm \hangafter=1 \noindent 
\textit{Heart drive} (Pathophysiology, continuous) [n-range: 0\%, hp-range: 0\%, missing: 100\%]: Acute coronary event, Pheochromocytoma, Thyrotoxicosis, Ventricular pre-excitation. 
 
\hangindent=0.5cm \hangafter=1 \noindent 
\textit{Heart post-load} (Pathophysiology, continuous) [n-range: 0\%, hp-range: 0\%, missing: 100\%]: Acute aortic valve failure, Acute mitral valve failure, Obstructive cardiomyopathy. 
 
\hangindent=0.5cm \hangafter=1 \noindent 
\textit{Heart rate (bpm)} (Semiotics (other), continuous) [lp-range: 4.71\%, n-range: 23.96\%, hp-range: 71.19\%, missing: 0.14\%]: Autonomic nervous system status, Bradycardia/Tachycardia. 
 
\hangindent=0.5cm \hangafter=1 \noindent 
\textit{Heartburn} (Semiotics (other), binary) [absent: 88.92\%, present: 7.06\%, missing: 4.02\%]: Dyspepsia. 
 
\hangindent=0.5cm \hangafter=1 \noindent 
\textit{Hemoglobin (gr/100 ml)} (Semiotics (other), continuous) [lp-range: 25.9\%, n-range: 72.85\%, missing: 1.25\%]: Acute anemia, Chronic anemia. 
 
\hangindent=0.5cm \hangafter=1 \noindent 
\textit{Hemopericardium} (Pathophysiology, binary) [absent: 0\%, present: 0\%, missing: 100\%]: Acute myocardial infarction, Aortic dissection. 
 
\hangindent=0.5cm \hangafter=1 \noindent 
\textit{Hemoptysis} (Semiotics (other), binary) [absent: 99.17\%, present: 0.83\%, missing: 0\%]: Lung cancer, Pneumonia, Pulmonary infarction, Upper airways infection. 
 
\hangindent=0.5cm \hangafter=1 \noindent 
\textit{Hemorrhage} (Pathology, binary) [absent: 0\%, present: 0.69\%, missing: 99.31\%]: Pancreatitis. 
 
\hangindent=0.5cm \hangafter=1 \noindent 
\textit{Hepatomegaly} (Semiotics (other), binary) [absent: 82.83\%, present: 9.97\%, missing: 7.2\%]: Right heart failure. 
 
\hangindent=0.5cm \hangafter=1 \noindent 
\textit{Herpes Zooster} (Pathology, binary) [absent: 99.31\%, present: 0.28\%, missing: 0.42\%]: - 
 
\hangindent=0.5cm \hangafter=1 \noindent 
\textit{Hiatal hernia} (Epidemiology, binary) [absent: 0\%, present: 0\%, missing: 100\%]: - 
 
\hangindent=0.5cm \hangafter=1 \noindent 
\textit{Hilar adenopathy} (Semiotics (other), binary) [absent: 0\%, present: 0\%, missing: 100\%]: Chronic interstitial lung disease, Lung cancer, Pneumonia. 
 
\hangindent=0.5cm \hangafter=1 \noindent 
\textit{Hyperhomocysteinemia} (Semiotics (other), binary) [absent: 2.08\%, present: 0.97\%, missing: 96.95\%]: Thrombophilia. 
 
\hangindent=0.5cm \hangafter=1 \noindent 
\textit{Hypertransparency} (Semiotics (other), multi-valued) [absent: 74.24\%, parenchymal: 5.4\%, pleuritic: 3.19\%, missing: 17.17\%]: Pulmonary emphysema, Spontaneous pneumothorax. 
 
\hangindent=0.5cm \hangafter=1 \noindent 
\textit{Hypoglycemia} (Semiotics (other), binary) [absent: 96.95\%, present: 1.94\%, missing: 1.11\%]: - 
 
\hangindent=0.5cm \hangafter=1 \noindent 
\textit{Iliac phlebography } (Semiotics (other), binary) [negative: 0.14\%, positive: 0\%, missing: 99.86\%]: Lower limbs deep vein thrombosis. 
 
\hangindent=0.5cm \hangafter=1 \noindent 
\textit{Immobilisation} (Semiotics (other), binary) [no: 85.32\%, yes: 14.68\%, missing: 0\%]: Neuromuscular disease. 
 
\hangindent=0.5cm \hangafter=1 \noindent 
\textit{Immunocompromission} (Pathophysiology, binary) [absent: 51.52\%, present: 0.83\%, missing: 47.65\%]: Age (years old), Neoplastic disease (generic). 
 
\hangindent=0.5cm \hangafter=1 \noindent 
\textit{Inspired oxygen fraction (percentage)} (Epidemiology, continuous) [air: 71.05\%, 0.21-0.35: 0\%, 0.35-0.50: 0\%, >0.50: 0\%, missing: 28.95\%]: - 
 
\hangindent=0.5cm \hangafter=1 \noindent 
\textit{Intra-vascular coagulation} (Pathogenesis, multi-valued) [normal: 0\%, ipernormal: 0\%, iponormal: 0\%, missing: 100\%]: Extrogens use, Neoplastic disease (generic), Obesity (Body Mass Index>=30), Prophylaxis/anticoagulation, Sepsis, Thrombophilia. 
 
\hangindent=0.5cm \hangafter=1 \noindent 
\textit{Jugular venous distention} (Semiotics (other), binary) [absent: 81.72\%, present: 5.54\%, missing: 12.74\%]: Right heart failure. 
 
\hangindent=0.5cm \hangafter=1 \noindent 
\textit{L-dopa use} (Epidemiology, binary) [no: 97.65\%, yes: 1.8\%, missing: 0.55\%]: - 
 
\hangindent=0.5cm \hangafter=1 \noindent 
\textit{Lactates (mmol/l)} (Semiotics (other), continuous) [n-range: 0\%, hp-range: 0\%, missing: 100\%]: Left cardiac output. 
 
\hangindent=0.5cm \hangafter=1 \noindent 
\textit{LDH} (Semiotics (other), binary) [normal: 52.91\%, augmented: 18.7\%, missing: 28.39\%]: Acute cerebro-vascular disease, Acute myocardial infarction, Bacterial infection, Neoplastic disease (generic), Pulmonary infarction, Thyroid disease. 
 
\hangindent=0.5cm \hangafter=1 \noindent 
\textit{Left bundle branch block} (Semiotics (other), binary) [absent: 87.4\%, present: 8.59\%, missing: 4.02\%]: Acute coronary event, Chronic cardiac muscle  disease. 
 
\hangindent=0.5cm \hangafter=1 \noindent 
\textit{Left cardiac output} (Pathophysiology, continuous) [lp-range: 0\%, n-range: 0\%, hp-range: 0\%, missing: 100\%]: Heart drive, Left heart pump, Left ventricular pre-load. 
 
\hangindent=0.5cm \hangafter=1 \noindent 
\textit{Left heart pump} (Pathophysiology, continuous) [lp-range: 0\%, n-range: 0\%, missing: 100\%]: Acute myocardial infarction, Chronic cardiac muscle  disease, Heart drive, Heart post-load. 
 
\hangindent=0.5cm \hangafter=1 \noindent 
\textit{Left ventricular hypertrophy} (Pathogenesis, binary) [absent: 56.09\%, present: 0.14\%, missing: 43.77\%]: Age (years old), Chronic aortic valve failure, Chronic arterial hypertension, Chronic mitral valve failure. 
 
\hangindent=0.5cm \hangafter=1 \noindent 
\textit{Left ventricular pre-load} (Pathophysiology, continuous) [lp-range: 0\%, n-range: 0\%, hp-range: 0\%, missing: 100\%]: Cardiac tamponade, Heart drive, Left heart pump, Right heart output. 
 
\hangindent=0.5cm \hangafter=1 \noindent 
\textit{Left ventricular thickness (>=5 mm)} (Semiotics (other), binary) [absent: 0\%, present: 0\%, missing: 100\%]: Cor pulmonale. 
 
\hangindent=0.5cm \hangafter=1 \noindent 
\textit{Leiden factor V} (Semiotics (other), binary) [absent: 1.94\%, present: 0.28\%, missing: 97.78\%]: Thrombophilia. 
 
\hangindent=0.5cm \hangafter=1 \noindent 
\textit{Leukemia} (Epidemiology, binary) [absent: 0\%, present: 0\%, missing: 100\%]: Neoplastic disease (generic). 
 
\hangindent=0.5cm \hangafter=1 \noindent 
\textit{Leukemic blast brisis} (Pathophysiology, binary) [absent: 99.17\%, present: 0.83\%, missing: 0\%]: Leukemia. 
 
\hangindent=0.5cm \hangafter=1 \noindent 
\textit{Leukocytosis} (Semiotics (other), multi-valued) [absent: 76.45\%, hypo: 19.94\%, hyper (moderate): 1.8\%, hyper (severe): 0.55\%, missing: 1.25\%]: Focal neurological signs, Lymphocytosis. 
 
\hangindent=0.5cm \hangafter=1 \noindent 
\textit{Lower limbs compression ultrasounds} (Semiotics (other), binary) [negative: 2.91\%, positive: 1.8\%, missing: 95.29\%]: Lower limbs deep vein thrombosis. 
 
\hangindent=0.5cm \hangafter=1 \noindent 
\textit{Lower limbs deep vein thrombosis} (Pathology, binary) [absent: 0\%, present: 0\%, missing: 100\%]: Venous intra-vascular coagulation. 
 
\hangindent=0.5cm \hangafter=1 \noindent 
\textit{Lower limbs echo-color doppler} (Semiotics (other), binary) [negative: 8.31\%, positive: 5.12\%, missing: 86.57\%]: Lower limbs deep vein thrombosis. 
 
\hangindent=0.5cm \hangafter=1 \noindent 
\textit{Lower limbs fractures} (Semiotics (other), binary) [absent: 97.65\%, present: 2.35\%, missing: 0\%]: - 
 
\hangindent=0.5cm \hangafter=1 \noindent 
\textit{Lower limbs magnetic resonance phlebography} (Semiotics (other), binary) [negative: 0\%, positive: 0\%, missing: 100\%]: Lower limbs deep vein thrombosis. 
 
\hangindent=0.5cm \hangafter=1 \noindent 
\textit{Lower limbs pain} (Semiotics (future), binary) [absent: 87.67\%, present: 6.23\%, missing: 6.09\%]: Lower limbs deep vein thrombosis. 
 
\hangindent=0.5cm \hangafter=1 \noindent 
\textit{Lung cancer} (Pathology, binary) [absent: 85.46\%, present: 3.88\%, missing: 10.66\%]: Neoplastic disease (generic). 
 
\hangindent=0.5cm \hangafter=1 \noindent 
\textit{Lung perfusion} (Pathophysiology, continuous) [n-range: 0\%, hp-range: 0\%, missing: 100\%]: Pulmonary hypertension, Pulmonary venous thrombo-embolism. 
 
\hangindent=0.5cm \hangafter=1 \noindent 
\textit{Lung perfusion scintigraphy} (Semiotics (other), binary) [negative: 0.97\%, positive: 2.35\%, missing: 96.68\%]: Bias of perfusion scintigraphy, Lung perfusion. 
 
\hangindent=0.5cm \hangafter=1 \noindent 
\textit{Lymphocytosis} (Semiotics (other), binary) [absent: 39.47\%, present: 8.86\%, missing: 51.66\%]: Non-bacterial infection. 
 
\hangindent=0.5cm \hangafter=1 \noindent 
\textit{Mallory-Weiss syndrome} (Pathology, binary) [absent: 0\%, present: 0\%, missing: 100\%]: Alcoholism. 
 
\hangindent=0.5cm \hangafter=1 \noindent 
\textit{Miller index} (Semiotics (other), multi-valued) [less (than 1): 0\%, tra (1 e 16): 0.14\%, more (than 16): 0\%, missing: 99.86\%]: Pulmonary venous thrombo-embolism. 
 
\hangindent=0.5cm \hangafter=1 \noindent 
\textit{Minute ventilation} (Pathophysiology, continuous) [lp-range: 0\%, n-range: 0\%, hp-range: 0\%, missing: 100\%]: Abnormal ventilation trigger, Acute anemia, Acute pulmonary disease, Anxiety/agitation, Lactates (mmol/l), Neuromuscular disease, Sepsis. 
 
\hangindent=0.5cm \hangafter=1 \noindent 
\textit{Mitral valve failure} (Semiotics (other), binary) [absent: 17.59\%, present: 24.52\%, missing: 57.89\%]: Acute mitral valve failure, Chronic mitral valve failure. 
 
\hangindent=0.5cm \hangafter=1 \noindent 
\textit{Mitral valve prolapse (generic)} (Semiotics (other), binary) [absent: 38.92\%, present: 2.63\%, missing: 58.45\%]: Acute mitral valve prolapse, Chronic mitral valve prolapse. 
 
\hangindent=0.5cm \hangafter=1 \noindent 
\textit{Myocardial stretching} (Pathophysiology, binary) [absent: 0\%, present: 0\%, missing: 100\%]: Acute aortic valve failure, Left ventricular pre-load, Pulmonary venous thrombo-embolism. 
 
\hangindent=0.5cm \hangafter=1 \noindent 
\textit{Myocarditis} (Pathology, binary) [absent: 24.52\%, present: 0.97\%, missing: 74.52\%]: Endocarditis, Non-infarctual pericarditis, Sepsis. 
 
\hangindent=0.5cm \hangafter=1 \noindent 
\textit{Myoglobin} (Semiotics (other), binary) [normal: 14.96\%, abnormal: 11.91\%, missing: 73.13\%]: Acute myocardial infarction. 
 
\hangindent=0.5cm \hangafter=1 \noindent 
\textit{Nausea} (Semiotics (other), binary) [absent: 82.13\%, present: 12.47\%, missing: 5.4\%]: Dyspepsia. 
 
\hangindent=0.5cm \hangafter=1 \noindent 
\textit{Neoplastic disease (generic)} (Semiotics (other), binary) [absent: 84.07\%, present: 14.96\%, missing: 0.97\%]: Age (years old), Gender, Smoker. 
 
\hangindent=0.5cm \hangafter=1 \noindent 
\textit{Neuromuscular disease} (Epidemiology, binary) [absent: 85.32\%, present: 0.83\%, missing: 13.85\%]: - 
 
\hangindent=0.5cm \hangafter=1 \noindent 
\textit{Nodule} (Semiotics (other), binary) [absent: 87.4\%, present: 7.48\%, missing: 5.12\%]: Lung cancer. 
 
\hangindent=0.5cm \hangafter=1 \noindent 
\textit{Non-bacterial infection} (Pathophysiology, binary) [absent: 0\%, present: 0\%, missing: 100\%]: Myocarditis, Non-infarctual pericarditis, Pancreatitis, Pulmonary infarction, Upper airways infection. 
 
\hangindent=0.5cm \hangafter=1 \noindent 
\textit{Non-infarctual pericarditis} (Pathology, binary) [absent: 92.66\%, present: 0.69\%, missing: 6.65\%]: - 
 
\hangindent=0.5cm \hangafter=1 \noindent 
\textit{Non-infective pericarditis} (Pathology, binary) [absent: 61.91\%, present: 1.52\%, missing: 36.57\%]: Acute myocardial infarction. 
 
\hangindent=0.5cm \hangafter=1 \noindent 
\textit{Non ST segment elevation} (Semiotics (other), multi-valued) [absent: 75.21\%, depressed (ST): 4.99\%, negative (T waves): 9.97\%, missing: 9.83\%]: Acute coronary event, Chronic cardiac muscle  disease, Left bundle branch block, Myocarditis. 
 
\hangindent=0.5cm \hangafter=1 \noindent 
\textit{Obesity (Body Mass Index>=30)} (Aetiology, binary) [BMI (less than 30): 96.4\%, BMI (30 or more): 3.6\%, missing: 0\%]: - 
 
\hangindent=0.5cm \hangafter=1 \noindent 
\textit{Obstruction of the systemic circulation} (Pathophysiology, binary) [absent: 0\%, present: 0\%, missing: 100\%]: Lung perfusion, Obstructive cardiomyopathy. 
 
\hangindent=0.5cm \hangafter=1 \noindent 
\textit{Obstructive cardiomyopathy} (Pathology, multi-valued) [absent: 46.26\%, stadium (I): 0.42\%, stadium (II): 0\%, missing: 53.32\%]: Age (years old), Aortic stenosis, Left ventricular hypertrophy. 
 
\hangindent=0.5cm \hangafter=1 \noindent 
\textit{Oliguria/anuria} (Semiotics (other), binary) [absent: 88.64\%, present: 6.37\%, missing: 4.99\%]: Left cardiac output. 
 
\hangindent=0.5cm \hangafter=1 \noindent 
\textit{Opacity to chest X-rays} (Semiotics (other), multi-valued) [absent: 68.01\%, nodular (reticulum): 18.01\%, widespread: 6.09\%, missing: 7.89\%]: Pulmonary opacity. 
 
\hangindent=0.5cm \hangafter=1 \noindent 
\textit{Orthopnea} (Semiotics (other), binary) [absent: 75.76\%, present: 13.99\%, missing: 10.25\%]: Chronic cardiac muscle  disease, Pulmonary edema, Pulmonary emphysema. 
 
\hangindent=0.5cm \hangafter=1 \noindent 
\textit{Orthostatic hypotension} (Semiotics (other), binary) [absent: 53.46\%, present: 1.94\%, missing: 44.6\%]: Dehydration, L-dopa use, Psychiatric medication, Pulmonary venous thrombo-embolism. 
 
\hangindent=0.5cm \hangafter=1 \noindent 
\textit{Oxygen saturation (percentage)} (Semiotics (other), continuous) [lp-range: 64.96\%, n-range: 22.3\%, missing: 12.74\%]: Inspired oxygen fraction (percentage), Lung perfusion, Minute ventilation, Pulmonary shunt. 
 
\hangindent=0.5cm \hangafter=1 \noindent 
\textit{paCO2 (mmHg)} (Semiotics (other), continuous) [lp-range: 18.56\%, n-range: 21.75\%, hp-range: 8.73\%, missing: 50.97\%]: Lung perfusion, Minute ventilation, Pulmonary shunt. 
 
\hangindent=0.5cm \hangafter=1 \noindent 
\textit{Palpitations} (Semiotics (future), binary) [absent: 71.19\%, present: 23.27\%, missing: 5.54\%]: Extrasystoles, Heart rate (bpm). 
 
\hangindent=0.5cm \hangafter=1 \noindent 
\textit{Pancreatitis} (Pathology, binary) [absent: 90.3\%, present: 0.14\%, missing: 9.56\%]: Age (years old), Alcoholism, Cholelithiasis, Gender. 
 
\hangindent=0.5cm \hangafter=1 \noindent 
\textit{paO2 (mmHg)} (Semiotics (other), continuous) [lp-range: 33.93\%, n-range: 14.27\%, missing: 51.8\%]: Oxygen saturation (percentage). 
 
\hangindent=0.5cm \hangafter=1 \noindent 
\textit{Paradoxical interventricular septum} (Semiotics (other), binary) [absent: 30.06\%, present: 1.39\%, missing: 68.56\%]: Right heart pre-load. 
 
\hangindent=0.5cm \hangafter=1 \noindent 
\textit{Patent foramen ovale} (Semiotics (other), binary) [normal: 41.14\%, pervious: 0.97\%, missing: 57.89\%]: - 
 
\hangindent=0.5cm \hangafter=1 \noindent 
\textit{Peptic ulcer} (Pathology, binary) [absent: 88.92\%, present: 0.28\%, missing: 10.8\%]: - 
 
\hangindent=0.5cm \hangafter=1 \noindent 
\textit{Pericardial effusion} (Semiotics (other), binary) [absent: 48.75\%, present: 3.05\%, missing: 48.2\%]: Hemopericardium, Pericarditis, Right heart failure. 
 
\hangindent=0.5cm \hangafter=1 \noindent 
\textit{Pericarditis} (Pathology, binary) [absent: 0\%, present: 0\%, missing: 100\%]: Non-infarctual pericarditis, Non-infective pericarditis. 
 
\hangindent=0.5cm \hangafter=1 \noindent 
\textit{Peripheral edema} (Semiotics (other), multi-valued) [absent: 77.98\%, unilateral: 4.29\%, bilateral: 16.62\%, missing: 1.11\%]: Chronic cardiac muscle  disease, Lower limbs deep vein thrombosis, Lower limbs fractures, Right heart failure. 
 
\hangindent=0.5cm \hangafter=1 \noindent 
\textit{Peritonitis} (Pathology, binary) [absent: 82.55\%, present: 0\%, missing: 17.45\%]: Cholecystitis, Peptic ulcer. 
 
\hangindent=0.5cm \hangafter=1 \noindent 
\textit{pH} (Semiotics (other), continuous) [lp-range: 6.09\%, n-range: 22.71\%, hp-range: 20.64\%, missing: 50.55\%]: Chronic metabolic alkalosis, Lactates (mmol/l), paCO2 (mmHg). 
 
\hangindent=0.5cm \hangafter=1 \noindent 
\textit{Pheochromocytoma} (Pathology, binary) [absent: 0\%, present: 0\%, missing: 100\%]: Age (years old). 
 
\hangindent=0.5cm \hangafter=1 \noindent 
\textit{Pleural effusion} (Semiotics (other), multi-valued) [absent: 76.59\%, unilateral: 12.74\%, bilateral: 5.54\%, missing: 5.12\%]: Lung cancer, Pleurisy, Pulmonary infarction, Right heart failure. 
 
\hangindent=0.5cm \hangafter=1 \noindent 
\textit{Pleurisy} (Pathology, binary) [absent: 18.28\%, present: 0.83\%, missing: 80.89\%]: Lung cancer, Non-infarctual pericarditis, Non-infective pericarditis, Pneumonia, Pulmonary infarction. 
 
\hangindent=0.5cm \hangafter=1 \noindent 
\textit{Pneumonia} (Pathology, multi-valued) [absent: 58.73\%, interstiziale: 1.52\%, solida: 0.69\%, missing: 39.06\%]: Age (years old), Alcoholism, Chronic cardiac muscle  disease, Chronic cerebro-vascular disease, Gender, Immunocompromission, Lung cancer, Pulmonary emphysema. 
 
\hangindent=0.5cm \hangafter=1 \noindent 
\textit{Pregnancy} (Aetiology, multi-valued) [no: 99.31\%, prepartum: 0\%, postpartum: 0.69\%, missing: 0\%]: Extrogens use, Fertility. 
 
\hangindent=0.5cm \hangafter=1 \noindent 
\textit{Previous episode of deep venous thrombosis/pulmonary embolism} (Aetiology, binary) [absent: 92.94\%, present: 7.06\%, missing: 0\%]: Thrombophilia. 
 
\hangindent=0.5cm \hangafter=1 \noindent 
\textit{Previous transient seizure} (Pathophysiology, multi-valued) [absent: 0\%, recently occurred: 0\%, present: 0\%, missing: 100\%]: Cerebral hypoxia, Cerebral mass, Hypoglycemia. 
 
\hangindent=0.5cm \hangafter=1 \noindent 
\textit{Prophylaxis/anticoagulation} (Epidemiology, multi-valued) [nessuna: 83.8\%, heparine: 2.49\%, anticoagulants: 5.12\%, missing: 8.59\%]: Chronic atrial arrhythmia, Previous episode of deep venous thrombosis/pulmonary embolism, Surgery. 
 
\hangindent=0.5cm \hangafter=1 \noindent 
\textit{Protein C} (Semiotics (other), binary) [normal: 2.35\%, deficit: 0\%, missing: 97.65\%]: Thrombophilia. 
 
\hangindent=0.5cm \hangafter=1 \noindent 
\textit{Protein S} (Semiotics (other), binary) [normal: 2.22\%, deficit: 0\%, missing: 97.78\%]: Thrombophilia. 
 
\hangindent=0.5cm \hangafter=1 \noindent 
\textit{Psychiatric medication} (Epidemiology, binary) [no: 89.61\%, yes: 10.39\%, missing: 0\%]: Anxiety/agitation. 
 
\hangindent=0.5cm \hangafter=1 \noindent 
\textit{Pulmonary artery diameter} (Semiotics (other), binary) [normal: 0\%, over (27 mm): 0\%, missing: 100\%]: Dilatated pulmonary artery disease. 
 
\hangindent=0.5cm \hangafter=1 \noindent 
\textit{Pulmonary artery thrombosis} (Semiotics (other), binary) [absent: 0\%, present: 0\%, missing: 100\%]: Dilatated pulmonary artery disease, Pulmonary venous thrombo-embolism. 
 
\hangindent=0.5cm \hangafter=1 \noindent 
\textit{Pulmonary consolidation} (Semiotics (other), multi-valued) [absent: 90.3\%, other (consolidation): 4.02\%, Hampton (hump): 0.28\%, missing: 5.4\%]: Lung cancer, Pneumonia, Pulmonary edema, Pulmonary infarction. 
 
\hangindent=0.5cm \hangafter=1 \noindent 
\textit{Pulmonary edema} (Pathology, multi-valued) [absent: 0\%, initial: 0\%, advanced: 0\%, missing: 100\%]: Acute respiratory distress syndrome, Left ventricular pre-load. 
 
\hangindent=0.5cm \hangafter=1 \noindent 
\textit{Pulmonary emphysema} (Epidemiology, multi-valued) [absent: 0\%, initial: 0\%, advanced: 0\%, missing: 100\%]: Age (years old), Gender, Smoker. 
 
\hangindent=0.5cm \hangafter=1 \noindent 
\textit{Pulmonary hypertension} (Pathophysiology, binary) [absent: 58.17\%, present: 0.69\%, missing: 41.14\%]: Previous episode of deep venous thrombosis/pulmonary embolism, Pulmonary emphysema. 
 
\hangindent=0.5cm \hangafter=1 \noindent 
\textit{Pulmonary infarction} (Pathophysiology, binary) [absent: 38.37\%, present: 2.63\%, missing: 59\%]: Pulmonary venous thrombo-embolism. 
 
\hangindent=0.5cm \hangafter=1 \noindent 
\textit{Pulmonary interstitium} (Pathophysiology, binary) [normal: 37.26\%, tickened: 29.5\%, missing: 33.24\%]: Pulmonary opacity. 
 
\hangindent=0.5cm \hangafter=1 \noindent 
\textit{Pulmonary opacity} (Pathophysiology, multi-valued) [absent: 0\%, interstitial: 0\%, alveolar: 0\%, missing: 100\%]: Age (years old), Chronic interstitial lung disease, Lung cancer, Pneumonia, Pulmonary edema, Pulmonary emphysema. 
 
\hangindent=0.5cm \hangafter=1 \noindent 
\textit{Pulmonary shunt} (Pathophysiology, continuous) [n-range: 0\%, hp-range: 0\%, missing: 100\%]: Acute pulmonary disease, Atelactasis, Pulmonary edema, Spontaneous pneumothorax. 
 
\hangindent=0.5cm \hangafter=1 \noindent 
\textit{Pulmonary venous thrombo-embolism} (Pathology, multi-valued) [absent: 0\%, non (massive): 0\%, massive: 0\%, missing: 100\%]: Lower limbs deep vein thrombosis, Right heart thrombus, Upper caval circle deep vein thrombosis. 
 
\hangindent=0.5cm \hangafter=1 \noindent 
\textit{Reflux of contrast medium into the hepatic veins} (Semiotics (other), binary) [absent: 11.77\%, present: 0.14\%, missing: 88.09\%]: Right heart failure. 
 
\hangindent=0.5cm \hangafter=1 \noindent 
\textit{Rib fracture} (Semiotics (other), binary) [absent: 49.31\%, present: 2.22\%, missing: 48.48\%]: Neoplastic disease (generic). 
 
\hangindent=0.5cm \hangafter=1 \noindent 
\textit{Right bundle branch block} (Semiotics (other), binary) [absent: 88.64\%, present: 10.11\%, missing: 1.25\%]: Cor pulmonale, Right heart pre-load. 
 
\hangindent=0.5cm \hangafter=1 \noindent 
\textit{Right circolatory obstruction trigger} (Pathophysiology, binary) [absent: 0\%, present: 0\%, missing: 100\%]: Pulmonary venous thrombo-embolism, Spontaneous pneumothorax. 
 
\hangindent=0.5cm \hangafter=1 \noindent 
\textit{Right heart failure} (Pathophysiology, continuous) [n-range: 0\%, hp-range: 0\%, missing: 100\%]: Cardiac tamponade, Left heart pump, Right heart pre-load. 
 
\hangindent=0.5cm \hangafter=1 \noindent 
\textit{Right heart output} (Pathophysiology, continuous) [lp-range: 0\%, n-range: 0\%, missing: 100\%]: Pulmonary venous thrombo-embolism, Right circolatory obstruction trigger, Right heart pre-load. 
 
\hangindent=0.5cm \hangafter=1 \noindent 
\textit{Right heart pre-load} (Pathophysiology, continuous) [lp-range: 0\%, n-range: 0\%, hp-range: 0\%, missing: 100\%]: Cor pulmonale, Dehydration, Pulmonary venous thrombo-embolism, Right circolatory obstruction trigger. 
 
\hangindent=0.5cm \hangafter=1 \noindent 
\textit{Right heart thrombus} (Semiotics (other), binary) [absent: 50.55\%, floating: 0\%, missing: 49.45\%]: Lower limbs deep vein thrombosis, Upper caval circle deep vein thrombosis, Venous intra-vascular coagulation. 
 
\hangindent=0.5cm \hangafter=1 \noindent 
\textit{Right ventricular hypokinesis} (Semiotics (other), binary) [absent: 41.69\%, present: 1.8\%, missing: 56.51\%]: Right heart failure. 
 
\hangindent=0.5cm \hangafter=1 \noindent 
\textit{Risk of deep vein thrombosis} (Pathogenesis, binary) [absent: 0\%, present: 0\%, missing: 100\%]: Chronic venous insufficiency, Lower limbs fractures, Neoplastic disease (generic), Smoker. 
 
\hangindent=0.5cm \hangafter=1 \noindent 
\textit{Ruptured chordae tendineae} (Semiotics (other), binary) [absent: 42.8\%, present: 0.14\%, missing: 57.06\%]: Acute mitral valve prolapse, Chronic mitral valve prolapse. 
 
\hangindent=0.5cm \hangafter=1 \noindent 
\textit{Sepsis} (Pathology, binary) [absent: 12.47\%, present: 0.69\%, missing: 86.84\%]: Peritonitis, Pneumonia. 
 
\hangindent=0.5cm \hangafter=1 \noindent 
\textit{Shock} (Semiotics (other), binary) [absent: 95.29\%, present: 1.8\%, missing: 2.91\%]: Lactates (mmol/l). 
 
\hangindent=0.5cm \hangafter=1 \noindent 
\textit{Sick sinus syndrome} (Pathophysiology, binary) [absent: 0\%, present: 0\%, missing: 100\%]: Age (years old). 
 
\hangindent=0.5cm \hangafter=1 \noindent 
\textit{Small pulmonary vessel diameter} (Semiotics (other), binary) [normal: 13.71\%, reduced: 1.39\%, missing: 84.9\%]: Dilatated pulmonary artery disease. 
 
\hangindent=0.5cm \hangafter=1 \noindent 
\textit{Smoker} (Aetiology, binary) [no: 90.58\%, yes: 9.42\%, missing: 0\%]: - 
 
\hangindent=0.5cm \hangafter=1 \noindent 
\textit{Sphincter incontinence} (Semiotics (other), binary) [absent: 96.68\%, present: 1.8\%, missing: 1.52\%]: Previous transient seizure. 
 
\hangindent=0.5cm \hangafter=1 \noindent 
\textit{Spontaneous pneumothorax} (Pathology, multi-valued) [absent: 90.58\%, limited: 1.66\%, extended: 1.52\%, missing: 6.23\%]: Age (years old), Lung cancer, Pulmonary emphysema. 
 
\hangindent=0.5cm \hangafter=1 \noindent 
\textit{ST segment elevation} (Semiotics (other), multi-valued) [absent: 85.18\%, non (ubiquitous ST): 4.02\%, ubiquitous (ST): 0.97\%, missing: 9.83\%]: Acute myocardial infarction, Left bundle branch block, Pericarditis. 
 
\hangindent=0.5cm \hangafter=1 \noindent 
\textit{Surgery} (Aetiology, multi-valued) [no: 96.68\%, general: 3.05\%, orthopedic: 0.28\%, missing: 0\%]: Lower limbs fractures. 
 
\hangindent=0.5cm \hangafter=1 \noindent 
\textit{Syncope} (Semiotics (future), binary) [absent: 82.96\%, present: 17.04\%, missing: 0\%]: Cerebral hypoxia, Hypoglycemia, Previous transient seizure. 
 
\hangindent=0.5cm \hangafter=1 \noindent 
\textit{T-wave inversion in V1-V3} (Semiotics (other), binary) [absent: 82.27\%, present: 7.89\%, missing: 9.83\%]: ECG right heart findings. 
 
\hangindent=0.5cm \hangafter=1 \noindent 
\textit{Tachypnea} (Semiotics (other), binary) [absent: 36.7\%, present: 40.86\%, missing: 22.44\%]: Minute ventilation. 
 
\hangindent=0.5cm \hangafter=1 \noindent 
\textit{Temporary suspension of heart drive} (Pathophysiology, binary) [absent: 82.96\%, present: 0.55\%, missing: 16.48\%]: Acute coronary event, Pheochromocytoma, Thyrotoxicosis, Ventricular pre-excitation. 
 
\hangindent=0.5cm \hangafter=1 \noindent 
\textit{Thrombophilia} (Pathogenesis, binary) [absent: 0\%, present: 0\%, missing: 100\%]: - 
 
\hangindent=0.5cm \hangafter=1 \noindent 
\textit{Thyroid-stimulating hormone (mUI/ml)} (Semiotics (other), continuous) [lp-range: 1.52\%, n-range: 25.48\%, hp-range: 1.66\%, missing: 71.33\%]: Thyroid disease. 
 
\hangindent=0.5cm \hangafter=1 \noindent 
\textit{Thyroid disease} (Epidemiology, multi-valued) [absent: 1.39\%, hypo (primitive): 0.14\%, hypo (secondary): 0.14\%, hyper (primitive): 0.55\%, hyper (secondary): 0.14\%, missing: 97.65\%]: Age (years old), Gender. 
 
\hangindent=0.5cm \hangafter=1 \noindent 
\textit{Thyroid hormones} (Pathophysiology, continuous) [lp-range: 0\%, n-range: 0\%, hp-range: 0\%, missing: 100\%]: Thyroid disease. 
 
\hangindent=0.5cm \hangafter=1 \noindent 
\textit{Thyrotoxicosis} (Pathology, binary) [absent: 55.54\%, present: 0.28\%, missing: 44.18\%]: Thyroid hormones. 
 
\hangindent=0.5cm \hangafter=1 \noindent 
\textit{Tongue bite} (Semiotics (other), binary) [absent: 80.75\%, present: 0.55\%, missing: 18.7\%]: Previous transient seizure. 
 
\hangindent=0.5cm \hangafter=1 \noindent 
\textit{Tricuspid valve insufficiency} (Semiotics (other), binary) [absent: 27.7\%, present: 12.33\%, missing: 59.97\%]: Chronic mitral valve failure, Cor pulmonale, Endocarditis, Right heart pre-load. 
 
\hangindent=0.5cm \hangafter=1 \noindent 
\textit{Troponin I} (Semiotics (other), binary) [normal: 50.69\%, augmented: 6.23\%, missing: 43.07\%]: Acute myocardial infarction, Myocardial stretching. 
 
\hangindent=0.5cm \hangafter=1 \noindent 
\textit{Upper airways infection} (Pathology, binary) [absent: 53.6\%, present: 8.59\%, missing: 37.81\%]: - 
 
\hangindent=0.5cm \hangafter=1 \noindent 
\textit{Upper caval circle deep vein thrombosis} (Pathology, binary) [absent: 0\%, present: 0\%, missing: 100\%]: Venous intra-vascular coagulation. 
 
\hangindent=0.5cm \hangafter=1 \noindent 
\textit{Urinary catecholamines} (Semiotics (other), binary) [negative: 0.69\%, positive: 0.14\%, missing: 99.17\%]: Pheochromocytoma. 
 
\hangindent=0.5cm \hangafter=1 \noindent 
\textit{Vasovagal syncope} (Pathology, binary) [absent: 82.96\%, present: 0.83\%, missing: 16.2\%]: - 
 
\hangindent=0.5cm \hangafter=1 \noindent 
\textit{Venous intra-vascular coagulation} (Pathogenesis, multi-valued) [absent: 17.73\%, upper (caval circle): 0.14\%, lower (caval circle): 10.53\%, right (heart): 0\%, missing: 71.61\%]: Central line, Chronic cardiac muscle  disease, Compression stockings, Immobilisation, Intra-vascular coagulation, Pregnancy, Previous episode of deep venous thrombosis/pulmonary embolism, Risk of deep vein thrombosis, Surgery. 
 
\hangindent=0.5cm \hangafter=1 \noindent 
\textit{Ventricular arrhythmia} (Semiotics (other), binary) [absent: 88.37\%, present: 2.08\%, missing: 9.56\%]: Bradycardia/Tachycardia. 
 
\hangindent=0.5cm \hangafter=1 \noindent 
\textit{Ventricular moderator band thickness} (Semiotics (other), binary) [absent: 26.04\%, present: 0.28\%, missing: 73.68\%]: Cor pulmonale. 
 
\hangindent=0.5cm \hangafter=1 \noindent 
\textit{Ventricular pre-excitation} (Aetiology, binary) [absent: 36.98\%, present: 0.55\%, missing: 62.47\%]: - 
 
\hangindent=0.5cm \hangafter=1 \noindent 
\textit{Ventricular segmental dyssynergia} (Semiotics (other), multi-valued) [absent: 32.13\%, hypokinesia (or diskinesia): 7.76\%, akinesia: 3.46\%, aneurysm: 0.28\%, missing: 56.37\%]: Acute coronary event, Acute myocardial infarction, Dilated cardiomyopathy, Ventricular arrhythmia. 
 
\hangindent=0.5cm \hangafter=1 \noindent 
\textit{Vomit} (Semiotics (other), binary) [absent: 90.44\%, present: 9.56\%, missing: 0\%]: Cerebral mass, Dyspepsia. 

\end{document}